\documentclass[10pt,twocolumn,letterpaper]{article}

\usepackage{iccv}
\usepackage{times}
\usepackage{epsfig}
\usepackage{graphicx}
\usepackage{comment}

\usepackage{enumitem}
\usepackage{algorithm}%
\usepackage{algorithmic}

\usepackage{booktabs} %
\usepackage{amsmath,amsfonts}
\usepackage{amssymb,amsopn}
\usepackage{bm} %
\usepackage{dsfont}
\usepackage[accsupp]{axessibility} %

\usepackage{authblk}

\def\eg{\textit{e.g}., } 
\def\ie{\textit{i.e}., }

\usepackage[pagebackref=true,breaklinks=true,letterpaper=true,colorlinks,bookmarks=false]{hyperref}

\iccvfinalcopy %

\def\iccvPaperID{9539} %

\ificcvfinal\fi

\begin{document}

\title{Continual Learning on Noisy Data Streams via Self-Purified Replay}

\author[1,2]{Chris Dongjoo Kim\thanks{Equal Contribution}}
\newcommand\CoAuthorMark{\footnotemark[\arabic{footnote}]} %
\author[2]{Jinseo Jeong\protect\CoAuthorMark}
\author[2]{Sangwoo Moon}
\author[2]{Gunhee Kim}

\affil[1]{NALBI Inc.}
\affil[2]{Department of Computer Science and Engineering, Seoul National University, South Korea}
\affil[ ]{\tt\small \{cdjkim, jinseo, sangwoo.moon\}@vl.snu.ac.kr gunhee@snu.ac.kr}

\maketitle
\ificcvfinal\thispagestyle{empty}\fi

\begin{abstract}
Continually learning in the real world must overcome many challenges, among which noisy labels are a common and inevitable issue. %
In this work, we present a replay-based continual learning framework that simultaneously addresses both catastrophic forgetting and noisy labels for the first time.
Our solution is based on two observations; (i) forgetting can be mitigated even with noisy labels via self-supervised learning, and (ii) the purity of the replay buffer is crucial.
Building on this regard, we propose two key components of our method: (i) a self-supervised replay technique named \textit{Self-Replay} which can circumvent erroneous training signals arising from noisy labeled data,
and (ii) the Self-Centered filter that maintains a purified replay buffer via centrality-based stochastic graph ensembles.
The empirical results on MNIST, CIFAR-10, CIFAR-100, %
and WebVision with real-world noise demonstrate that our framework can maintain a highly pure replay buffer amidst noisy streamed data while greatly outperforming the combinations of the state-of-the-art continual learning and noisy label learning methods.
The source code is available at \url{http://vision.snu.ac.kr/projects/SPR}
\end{abstract}

\section{Introduction}
\label{introduction}
The most natural form of input for an intelligent agent occurs sequentially.
Hence, the ability to continually learn from sequential data has gained much attention in recent machine learning research.
This problem is often coined as \textit{continual learning}, for which three representative approaches have been proposed~\cite{mccloskey89,ratcliff90,french99} including replay~\cite{lopez17,hayes19memory,aljundi19gradient,shin17,rolnick19,lesort19b}, regularization~\cite{kirkpatrick17ewc,zenke17,aljundi19selfless}, and expansion techniques~\cite{rusu16,yoon18}.

At the same time, learning from data riddled with noisy labels is an inevitable scenario that an intelligent agent must overcome. %
There have been multiple lines of work to learn amidst noisy labels such as loss regularization~\cite{wang19sce, zhang18nips,hendrycks18nips},
data re-weighting~\cite{ren18l2r,shen19icml}, label cleaning~\cite{pleiss20aum,lee18cleannet,ostyakov18eccv}, and training procedures~\cite{wei20jocor,jiang17icml}.

In this work, we aim to jointly tackle the problems of \textit{continual learning} and \textit{noisy label classification},
which to the best of our knowledge have not been studied in prior work. %
Noisy labels and continual learning are inevitable for real-world machine learning, as data comes in a stream possibly polluted with label inconsistency.
Hence, the two are bound to intersect; we believe exploring this intersection may glean evidence for promising research directions and hopefully shed light on the development of sustainable real-world machine learning algorithms.

We take on the replay-based approach to tackle continual learning since it has often shown superior results in terms of performance and memory efficiency even with simplicity.
Yet, we discover that replaying a noisy buffer intensifies the forgetting process due to the fallacious mapping of previously attained knowledge. %
Moreover, existing noisy label learning approaches show great limitations when coping within the online task-free setting~\cite{aljundi19mir, prabhu19gdumb, lee20iclr, kim20eccv}.
In their original forms, they assume that the whole dataset is given to purify the noise and thus are hampered by a small amount of data stored only in the replay buffer to either regularize, re-weight, or decide on its validity.

We begin by backtracking the root of the problem; if we naively store a sampled set of the noisy input stream into the replay buffer, it becomes riddled with noise, worsening the amount of forgetting.
Thus, we discover the key to success is maintaining a pure replay buffer, which is the major motive of our novel framework named \textit{Self-Purified Replay} (SPR).
At the heart of our framework is self-supervised learning~\cite{devlin19bert, chen20icml, he20cvpr, grill20nips}, which allows to circumvent the erroneous training signals arising from the incorrect pairs of data and labels.
Within the framework, we present our novel \textit{Self-Replay} and \textit{Self-Centered filter} that collectively cleanse noisy labeled data and continually learn from them.
The Self-Replay mitigates the noise intensified catastrophic forgetting, and the Self-Centered filter achieves a highly clean replay buffer %
even when restricted to a small portion of data at a time.

We outline the contributions of this work as follows.

\begin{enumerate}%
\item To the best of our knowledge, this is the first work to tackle \textit{noisy labeled continual learning}.
We discover noisy labels exacerbate catastrophic forgetting, and it is critical to filter out such noise from the input data stream before storing them in the replay buffer.
\item We introduce a novel replay-based framework named Self-Purified Replay (SPR), for noisy labeled continual learning.
SPR can not only maintain a clean replay buffer but also effectively mitigate catastrophic forgetting with a fixed parameter size.
\item We evaluate our approach on three synthetic noise benchmarks of MNIST~\cite{lecun98}, CIFAR-10~\cite{krizhevsky09}, CIFAR-100~\cite{krizhevsky09} and one real noise dataset of WebVision~\cite{li17arxiv}. Empirical results validate that SPR significantly outperforms many combinations of the state-of-the-art continual learning and noisy label learning methods. 
\end{enumerate}

\section{Problem Statement}
\label{sec:motivation}

\subsection{Noisy Labeled Continual Learning}
We consider the problem of  online task-free continual learning for classification where a sample $\{x_t, y_t\}$ enters at each time step $t$ in a non i.i.d manner without task labels.
While previous works~\cite{aljundi19gradient, prabhu19gdumb, lee20iclr} assume $\{x_t, y_t\}$ are correct (clean) samples, we allow the chance that a large portion of the data is falsely labeled.

\subsection{Motivation: Noise induced Amnesia}
We discover that if the data stream has noisy labels, it traumatically damages the continual learning model, analogous to \textit{retrograde amnesia}~\cite{squire81}, the inability to recall experience of the past.
We perform some preliminary experiments on a sequential version of symmetric noisy MNIST and CIFAR-10~\cite{lyu20iclr, wang19sce} using experience replay with the conventional reservoir sampling technique~\cite{riemer19iclr,zhang17er}.

\begin{figure}[t]
\begin{center}
\includegraphics[width=\columnwidth]{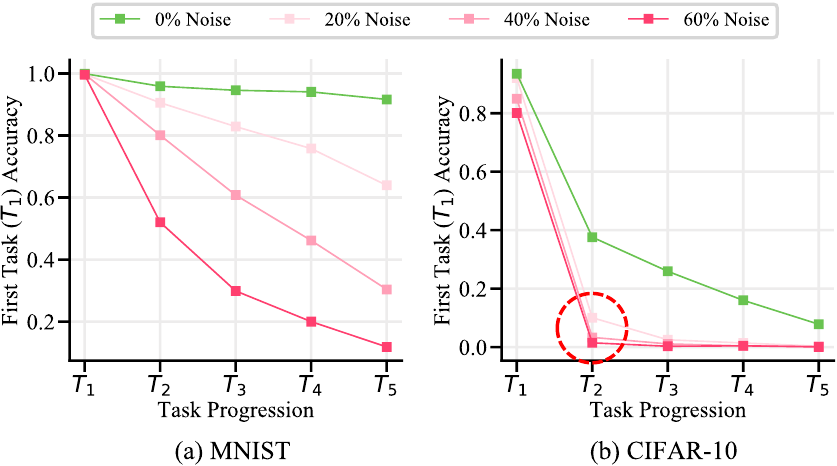}
\end{center}
\caption{A noisy labeled continual learning on the symmetric noisy in (a) MNIST~\cite{lecun98} and (b) CIFAR-10~\cite{krizhevsky09} when using experience replay with the conventional reservoir sampling~\cite{zhang17er, riemer19iclr}.
At the end of each task, the accuracy of the first task ($T_1$) is plotted. It shows that the noisy labels accelerate catastrophic forgetting. Notably, the dotted red circle in (b) indicates the significantly hastened forgetting process.}
\vspace{-3pt}
\label{fig:motivation}
\end{figure}

The empirical results in Figure~\ref{fig:motivation} show that when trained with noisy labels, the model becomes much more prone to catastrophic forgetting~\cite{french99, mccloskey89, thrun96, ratcliff90}.
As the noise level increases from 0\% to 60\%, sharp decreases in accuracy are seen.
Surprisingly, the dotted red circle in Figure~\ref{fig:motivation}(b) shows that in CIFAR-10 a fatally hastened forgetting occurs no matter the amount of noise.

We speculate that a critical issue that hinders the continual model is the corrupted replay buffer.
An ideal replay buffer should shield the model from noisy labels altogether by being vigilant of all the incoming data for the maintenance of a clean buffer.

\section{Approach to Noisy Labeled Continual Learning}
\label{sec:approach}
We design an approach to continual learning with noisy labels by realizing the two interrelated subgoals as follows.
\begin{enumerate}[start=1,label={\bfseries G\arabic*.}]
\item \textit{Reduce forgetting even with noisy labels}: The approach needs to mitigate catastrophic forgetting amidst learning from noisy labeled data. 

\item \textit{Filter clean data}: The method should learn representations such that it identifies the noise as anomalies.
Moreover, it should enable this from a \textit{small amount} of data since we do not have access to the entire dataset in online continual learning.
\end{enumerate}

\begin{figure}[t]
\begin{center}
\includegraphics[width=\columnwidth]{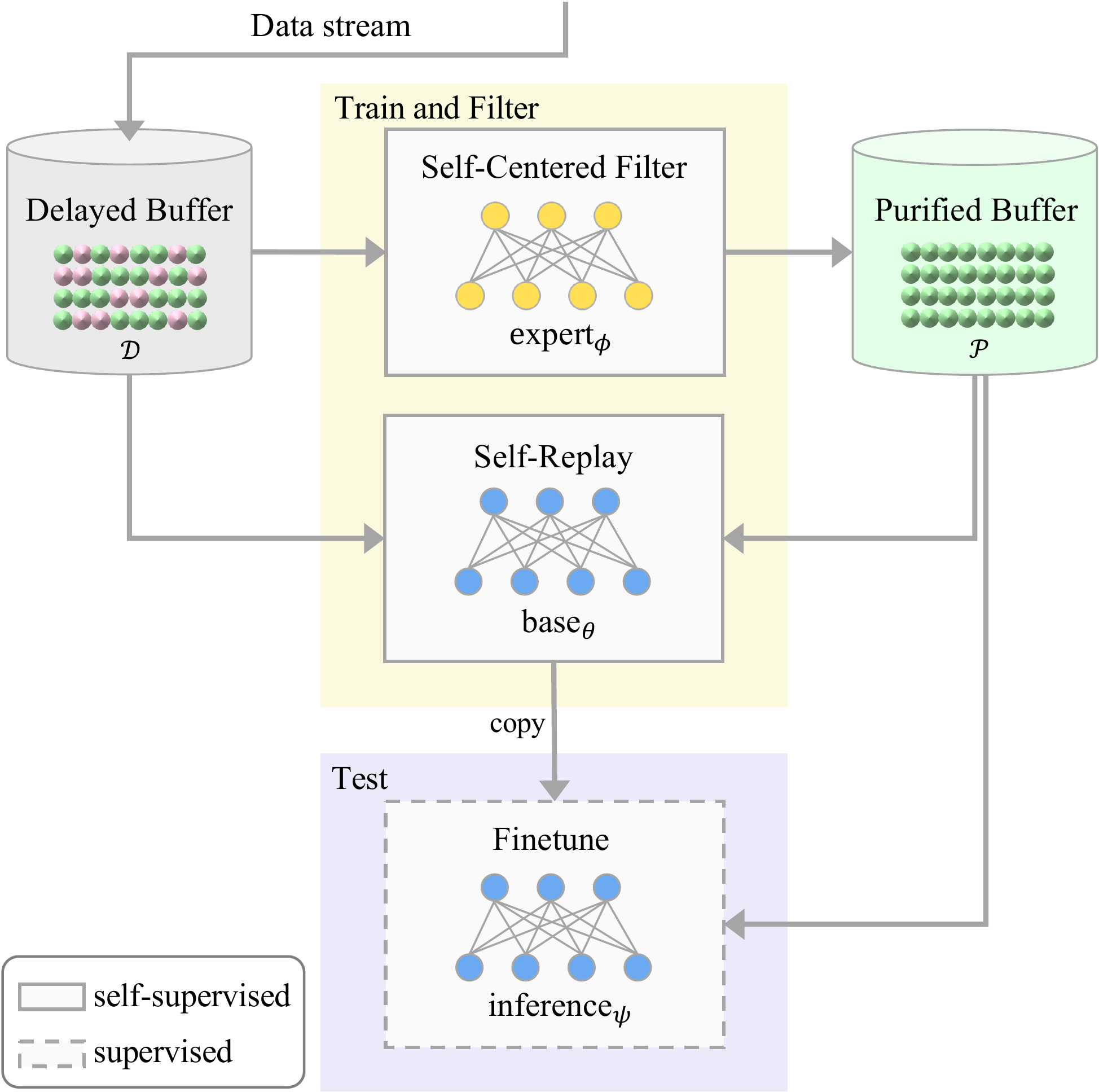}
\end{center}
\caption{Illustration of the Self-Purified Replay (SPR) framework. We specify the training and filtering phase (in the yellow shade) in Algorithm~\ref{alg:train_algo}, and the test phase (in the purple shade) in Algortihm~\ref{alg:eval_algo}.}
\vspace{-3pt}
\label{fig:framework}
\end{figure}

Figure~\ref{fig:framework} overviews the proposed framework consisting of two buffers and two networks. The \textit{delayed buffer}  $\mathcal{D}$  temporarily stocks the incoming data stream, and the \textit{purified buffer} $\mathcal{P}$ maintains the cleansed data.
The \textit{base} network addresses \textbf{G1} via self-supervised replay (Self-Replay) training (Section \ref{sec:self_replay}).
The \textit{expert} network is a key component of Self-Centered filter that tackles \textbf{G2} by obtaining confidently clean samples via centrality (Section \ref{sec:self_filtering}).
Both networks have the same architecture (\eg ResNet-18) with separate parameters.

Algorithm~\ref{alg:train_algo} outlines the training and filtering procedure.
Whenever the delayed buffer $\mathcal{D}$ is full, %
The Self-Centered filter powered by the expert network filters the clean samples from $\mathcal{D}$ to the purified buffer $\mathcal{P}$.
Then, the base network is trained via the self-supervision loss with the samples in $\mathcal{D} \cup \mathcal{P}$.
The detail will be discussed in Section \ref{sec:self_replay}--\ref{sec:self_filtering}.

At any stage of learning, we can perform downstream tasks (\ie classification) by duplicating the base network into the inference network, adding a final softmax layer, and finetuning it using the samples in $\mathcal{P}$.
Algorithm~\ref{alg:eval_algo} outlines this inference phase.

\subsection{Self-Replay}
\label{sec:self_replay}
Learning with noisy labeled data~\cite{pleiss20aum, arpit17, ma18, harutyunyan20} results in erroneous backpropagating signals when falsely paired $x$ and $y$ exist in the training set.
Hence, we circumvent this error via learning only from $x$ (without $y$) using contrastive self-supervised learning techniques~\cite{caron20nips, chen20icml, he20cvpr, grill20nips}.
That is, the framework first focuses on learning general representations via self-supervised learning from \textit{all} incoming $x$.
Subsequently, the downstream task (\ie supervised classification) finetunes the representation using only the samples in the purified buffer $\mathcal{P}$.
Building on this concept in terms of continual learning leads to Self-Replay, which mitigates forgetting while learning general representations via self-supervised replay of the samples in the delayed and purified buffer ($\mathcal{D} \cup \mathcal{P}$).

Specifically, we add a projection head $g(\cdot)$ (\ie a one-layer MLP) on top of the average pooling layer of the base network,
and train it using the normalized temperature-scaled cross-entropy loss~\cite{chen20icml}.
For a minibatch from $\mathcal{D}$ and $\mathcal{P}$ with a batch size of $B_d,B_p \in \mathbb{N}$ respectively,
we apply random image transformations (\eg cropping, color jitter, horizontal flip) to create two correlated views of each sample, referred to as positives.
Then, the loss is optimized to attract the features of the positives closer to each other while repelling them from the other samples in the batch, referred to as the negatives.
The updated objective becomes
\begin{align}
\label{eq:ntxent}
L_{self} =
- \hspace{-6pt}\sum_{i=1}^{2(B_d+B_p)} \hspace{-6pt}\text{log} \frac{e^{u_i^T u_j / \tau }}{\sum_{k=1}^{2(B_d+B_p)} \mathds{1}_{k \neq i} e^{u_i^T u_k / \tau}}. %
\end{align}
\noindent
We denote $(x_i, x_j)$ as the positives and $x_k$ as the negatives. $u_i = \frac{g(x_i)}{||g(x_i)||_2}$ is the $\ell_2$ normalized feature, and  $\tau > 0$ is the temperature.
Every time when the delayed buffer is full, we train the base network with this loss.

\begin{figure}[ht]
\begin{center}
\includegraphics[width=\columnwidth]{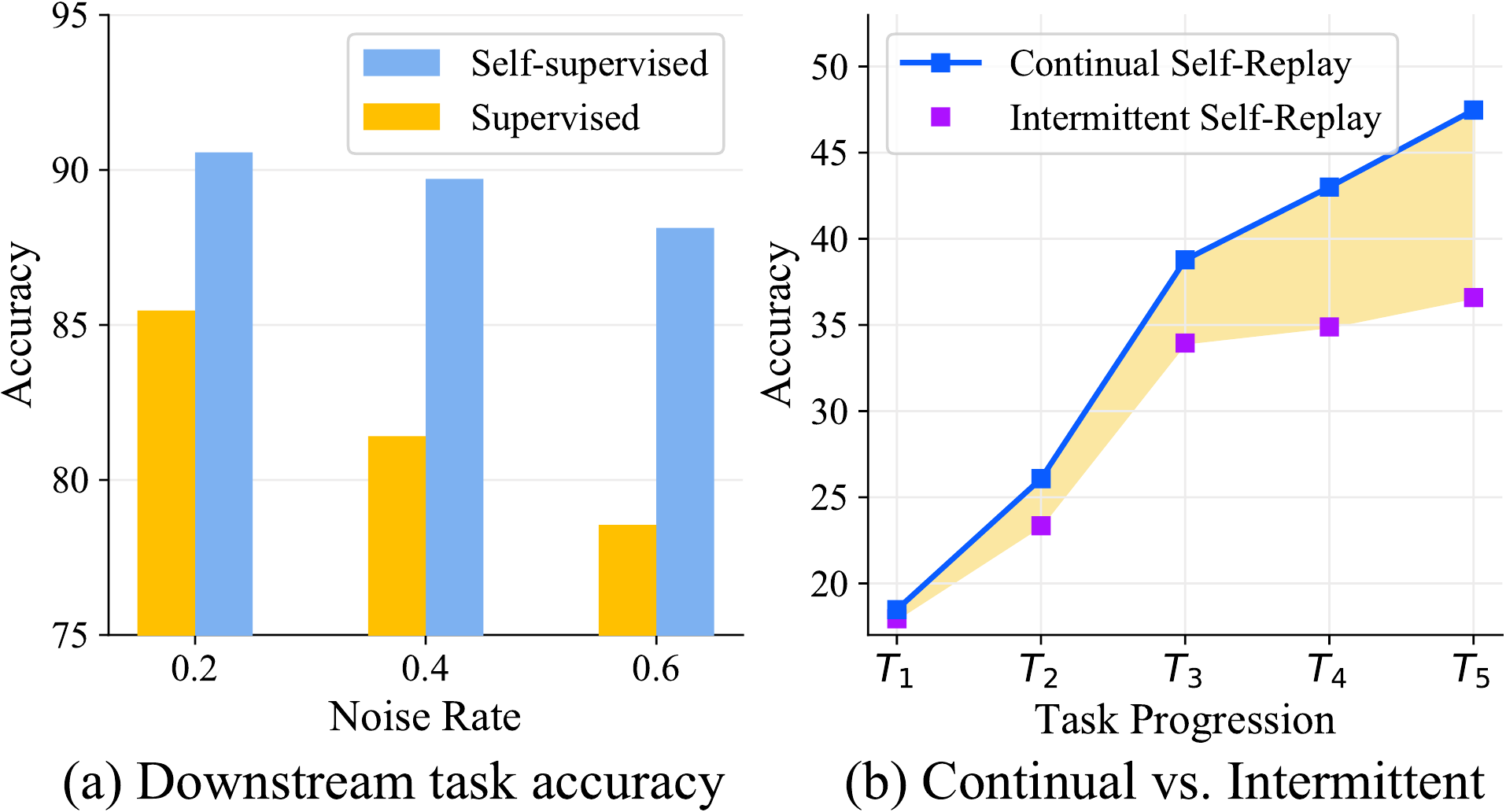}
\end{center}
\caption{\textbf{Empirical support for Self-Replay} with ResNet18 as the base network on CIFAR-10.
(a) Comparison of overall accuracy of the finetuned downstream classification between self-supervised and supervised representations trained on various noise rates. %
The self-supervised indicates that the base network trained using only $x$ as proposed, while the supervised means training with possibly noisy $(x,y)$ pairs. 
(b) The benefits of \textit{continual} Self-Replay over the \textit{intermittent} Self-Replay by comparing the test set accuracy of finetuned models.
The intermittent Self-Replay means training only with contents of the purified buffer up to and including the current task.}
\label{fig:self_replay_charac}
\end{figure}

\textbf{Empirical supports}. Figure~\ref{fig:self_replay_charac} shows some empirical results about the validity of Self-Replay for noisy labeled continual learning.
\begin{enumerate}[label=$\bullet$]

\item Figure~\ref{fig:self_replay_charac}(a) shows a quantitative examination on downstream classification tasks. It indicates that self-supervised learning leads to a better representation, and eventually outperforms the supervised one by noticeable margins.

\item Figure~\ref{fig:self_replay_charac}(b) exemplifies the superiority of Self-Replay in continual learning. 
  We contrast the performances of continually trained Self-Replay (as proposed) against intermittently trained Self-Replay, which trains \textit{offline} with only the samples in the purified buffer at the end of each task. 
  The colored areas in Figure~\ref{fig:self_replay_charac}(b) indicate how much the continually learned representations alleviate the forgetting and benefit the knowledge transfers among the past and future tasks. %
\end{enumerate}

\begin{algorithm}[tb]
    \caption{Training and filtering phase of SPR}
    \label{alg:train_algo}
    \begin{algorithmic}
        \STATE {\bfseries Input:} Training data $(x_t, y_t),...,(x_T, y_T)$ and initial parameters of base network $\theta$. %
        \STATE $\mathcal{D} = \mathcal{P} = \{\}$ \text{ // Initialize delayed and purified buffer}

        \FOR{$t=1$ {\bfseries to} $T$}
        \IF{$\mathcal{D}$ is full}
        \STATE $\mathcal{P} \leftarrow \mathcal{P}  \ \cup $ Self-Centered Filter($D$) \text{(section \ref{sec:self_filtering})}
        \STATE $\theta\leftarrow$ Self-Replay using $\mathcal{D} \cup \mathcal{P}$ \text{(section \ref{sec:self_replay})}
            \STATE reset $\mathcal{D}$
        \ELSE
        \STATE update $\mathcal{D}$ with $(x_t, y_t)$
        \ENDIF
        \ENDFOR
    \end{algorithmic}
\end{algorithm}

\begin{algorithm}[tb]
    \caption{Test phase of SPR}
    \label{alg:eval_algo}
    \begin{algorithmic}
        \STATE {\bfseries Input:} Test data $(x_t, y_t),...,(x_T, y_T)$,  parameters of the base network  $\theta$, and purified buffer $\mathcal{P}$
        \STATE $\psi$ = copy($\theta$) \text{// Duplicate base model to inference model}
        \STATE $\psi\leftarrow$ supervised finetune using $\mathcal{P}$
        \FOR{$t=1$ {\bfseries to} $T$}
        \STATE downstream classification  for $(x_t, y_t)$ using $\psi$
        \ENDFOR
    \end{algorithmic}
\end{algorithm}

\subsection{Self-Centered Filter}
\label{sec:self_filtering}

The goal of the Self-Centered filter is to obtain confidently clean samples; specifically, it assigns the probability of being clean to all the samples in the delayed buffer.

\textbf{Expert Network}. The expert network is prepared to \textit{featurize} the samples in the delayed buffer.
These features are used to compute the centrality of the samples, which is the yardstick of selecting clean samples. 
Inspired by the success of self-supervised learning good representations in Self-Replay, the expert network is also trained with the self-supervision loss in Eq.~\ref{eq:ntxent}
with only difference that we use the samples in $\mathcal{D}$ only  (instead of $\mathcal{D} \cup \mathcal{P}$ for the base network).

\textbf{Centrality}. At the core of the Self-Centered filter lies centrality~\cite{nieminen74}, which is rooted in graph theory to identify the most influential vertices within a graph.  %
We use a variant of the \textit{eigenvector centrality}~\cite{bonacich01}, %
which is grounded on the concept that a link to a highly influential vertex contributes to centrality more than a link to a lesser influential vertex.

First, weighted undirected graphs $G:=(V, E)$ are constructed \textit{per unique class label} in the delayed buffer.
We assume that the clean samples form the largest clusters in the graph of each class.
Each vertex $v \in V$ is a sample of the class, and the edge $e \in E$ is weighted by the cosine similarity between the features from the expert network.
For the adjacency matrix $\mathbf{A}= (a_{v,u})_{|V| \times |V|}$. Then the eigenvector centrality is formulated as
\begin{align}
  c_v = \frac{1}{\lambda} \sum_{u \in N(v)}c_u = \frac{1}{\lambda} \sum_{u \in V} a_{v,u} c_u, \label{eq:eigen_cent_vec} %
\end{align}
\noindent where $N(v)$ is the neighboring set of $v$, $\lambda$ is a constant and $a_{v,u}$ is the truncated similarity value within $(0, 1]$.
Eq.~\ref{eq:eigen_cent_vec} can be rewritten in vector notation as $\mathbf{A}\mathbf{c} = \lambda \mathbf{c}$, where $\mathbf{c}$ is a vectorized centrality over $V$.
The principal eigenvector $\mathbf{c}$ can be computed by the power method~\cite{vonMises1929}, and it corresponds to the eigenvector centrality for the vertices in $V$. %

\textbf{Beta Mixture Models}. The centrality quantifies which samples are the most influential (or the cleanest) within the data of identical class labels.
However, the identically labeled data contains both \textit{clean} and \textit{noisy} labeled samples, in which the noisy ones may deceptively manipulate the centrality score, leading to an indistinct division of the clean and noisy samples' centrality scores.
Hence, we compute the probability of cleanliness per sample via fitting a Beta mixture model (BMM)~\cite{jacobs1991MoE} to the centrality scores as
\begin{align}
p(c) = \sum_{z=1}^{Z} \pi_z p(c|z), \label{eq:bmm}
\end{align}
where $c > 0$ is the centrality score, $\pi_z$ is the mixing coefficients, and $Z\in\mathbb{N}$ is the number of components.
Beta distribution for $p(c|z)$  is a suitable choice due to the skewed nature of the centrality scores.
We set $Z=2$, indicating the clean and noisy components, and it is empirically the best in terms of accuracy and computation cost.
We use the EM algorithm~\cite{dempster77EM} to fit the BMM through which we obtain the posterior probability%
\begin{align}
 p(z|c) =  \frac{\pi_z p(c|\alpha_z, \beta_z)}{\sum_{j=1}^Z \pi_j p(c|\alpha_j, \beta_j)}, \label{eq:posterior}
\end{align}
where ${\alpha_z, \beta_z} > 0$ are the latent distribution parameters.
Please refer to the appendix for details of computing $ p(z|c)$.

Among the $Z=2$ components, we can easily identify the \textit{clean} component as the one that has the higher $c$ scores (\ie a larger cluster).
Then, the clean posterior $p(z=\mbox{clean}|c)$ defines the probability that centrality $c$ belongs to the clean component, which is used as the probability to enter and exit the purified buffer, $\mathcal{P}$.
After the selected samples enters our full purified buffer, the examples with the lowest $p(z=\mbox{clean}|c)$ are sampled out accordingly.

\begin{figure}[t]
\begin{center}
\includegraphics[width=\columnwidth]{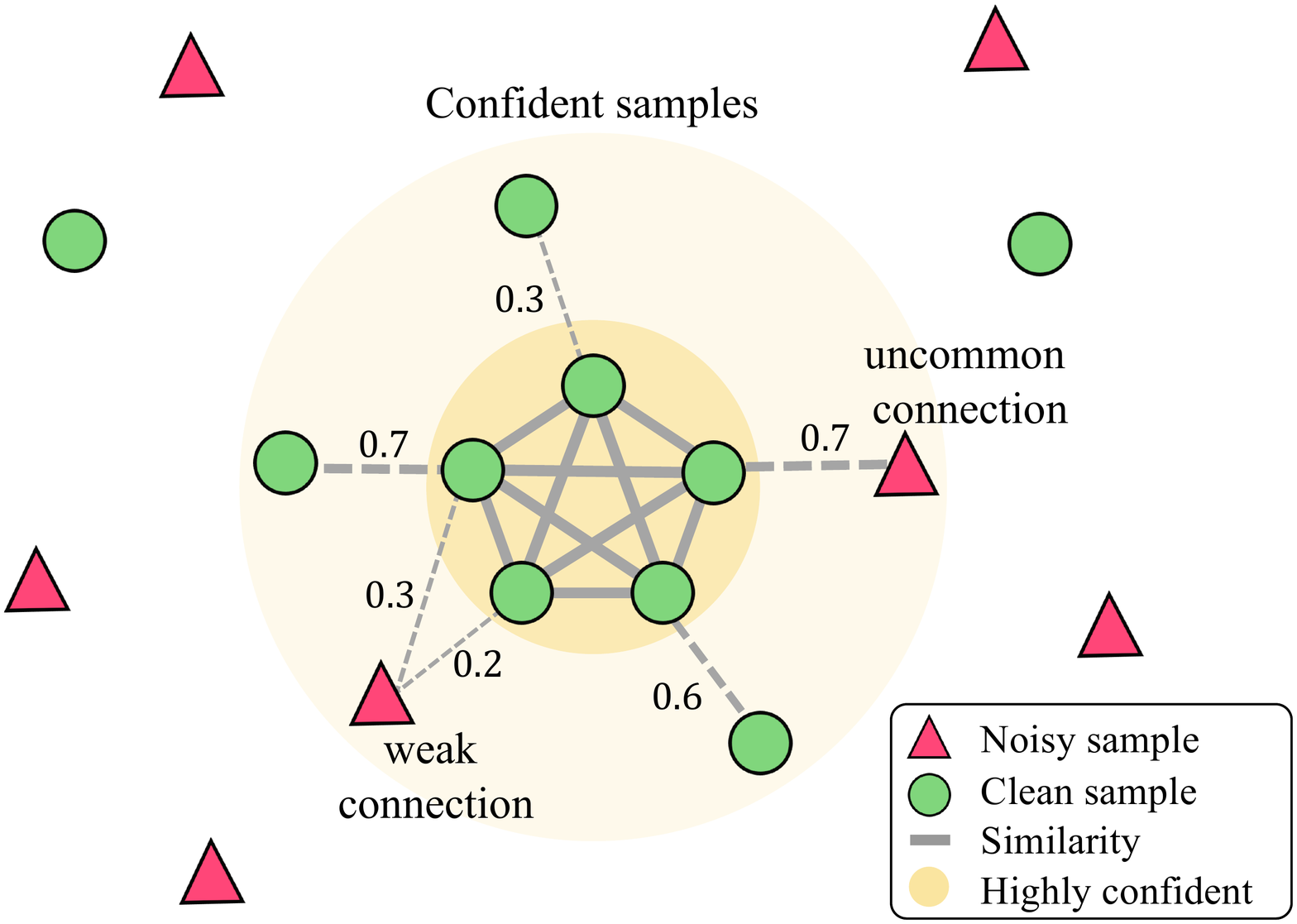}
\end{center}
\caption{Illustration of graph manipulation via Stochastic Ensemble, which severs weak and uncommon connections and probabilistically focus on confident and clean data within the graph.}
\label{fig:stochastic_ensemble_illust}
\end{figure}

\subsubsection{Stochastic Ensemble}
\label{stochastic_ensemble}
Since our goal is to obtain the \textit{most} clean samples as possible, we want to further sort out the possibly noisy samples.
We achieve this by introducing a \textit{stochastic ensemble} of BMMs, enabling a more noise robust posterior than the non-stochastic posterior $p(z=\text{clean}|c)$ in the previous section.

First, we prepare for stochastic ensembling by sampling multiple \textit{binary} adjacency matrices $\{ \mathbf{A} \}$ from a Bernoulli distribution over $\mathbf{A}$. 
For each class $l$, we impose a conditional Bernoulli distribution over $\mathbf{A}$ as
\begin{align}
    \hspace{-3pt} p(\mathbf{A}|D_{l}) =\hspace{-4pt} \prod_{d_i, d_j \in D_l} \hspace{-4pt} \text{Bern}\left(\mathbf{A}_{ij} | \text{ReLU}\left(\frac{d_{i} \cdot d_{j}}{||d_{i}|| ||d_{j}||}\right)\right),
\end{align}
where $D_l$ is the set of penultimate feature of class $l$ from the expert network.
We find that it is empirically helpful to truncate the dissimilar values to 0 (ReLU) and use the cosine similarity value as the probability. 
We replace the zeros in $\mathbf{A}$ with a small positive value to satisfy the requirement of Perron-Frobenius theorem\footnote{
  Perron-Frobenius theorem states when $\mathbf{A}$ has positive entries, it has a unique largest real eigenvalue, whose corresponding eigenvector have strictly positive components.}.
Then, our reformulated robust posterior probability is
\begin{align}
\label{eq:stochastic_posterior}
p(z|D_{l}) \propto \int_\mathbf{A} p(z|\text{cent}(\mathbf{A})) d p(\mathbf{A}|D_{l}), 
\end{align}
where $\text{cent}(\cdot)$ is the centrality scores from Eq.~\ref{eq:eigen_cent_vec}, and $p(z|\text{cent}(\mathbf{A}))$ can be obtained in the same manner as the \textit{non-stochastic} posterior in the previous section.
We approximate the integral using Monte Carlo sampling for which we use $E_{max}$ as the sample size.
Essentially, we fit the mixture models on different stochastic graphs to probabilistically carve out more confidently noisy samples by retaining the strong and dense connections while severing weak or uncommon connections.
This is conceptually illustrated in Figure~\ref{fig:stochastic_ensemble_illust}.

\begin{figure}[t!]
\begin{center}
\includegraphics[width=\columnwidth]{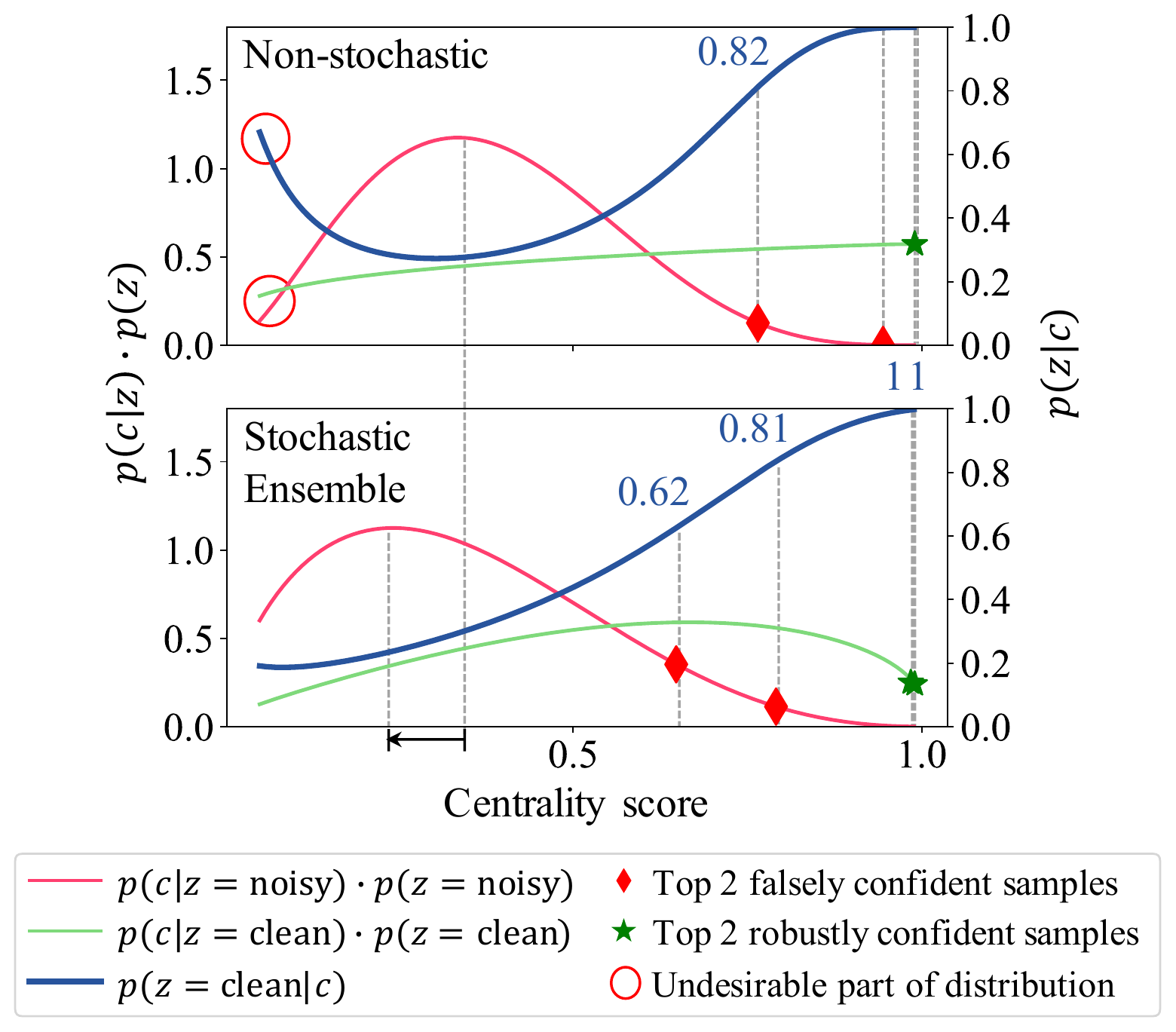}
\caption{Comparison of non-stochastic and Stochastic Ensemble on CIFAR-10 with 40\% noise. 
Stochastic Ensemble produces more confidently clean samples by shifting $p(c|z=\text{noisy}) \cdot p(z=\text{noisy})$ to the left,
and suppressing the cases where $p(c|z=\text{noisy})\cdot p(z=\text{noisy})$ dips below $p(c|z=\text{clean}) \cdot p(z=\text{clean})$.}
\label{fig:stochasticA_vs_fixedA}
\end{center}
\end{figure}

\textbf{Empirical Supports.} Figure~\ref{fig:stochasticA_vs_fixedA} shows some empirical evidence where the stochastic ensemble addresses the two issues to achieve a noise robust posterior $p(z|D_{l})$.
\begin{enumerate}[label=$\bullet$]
\item First, a small portion of noisy samples are falsely confident and are consequently assigned a high centrality score. Stochastic ensembling is able to suppress these noisy samples, as indicated in
Figure~\ref{fig:stochasticA_vs_fixedA}, where the mode of $p(c|z=\text{noisy})\cdot p(z=\text{noisy})$ (red curve) is shifted to the left by a noticeable margin.

\item Second, there are some cases where $p(c|z=\text{noisy})\cdot p(z=\text{noisy})$ drops below the $p(c|z=\text{clean})\cdot p(z=\text{clean})$ leading to a high $p(z=\text{clean}|c)$ for the noisy instances, indicated with red circles in Figure~\ref{fig:stochasticA_vs_fixedA}.
The stochastic ensemble of differing $\mathbf{A}$s can mitigate such problematic cases to drown out the unexpected noise.
\end{enumerate}

\section{Related Works}
\label{related_works}

\subsection{Continual Learning}
\label{continual_related_works}
There have been three main branches to train a model from continual data streams: regularization~\cite{li16lwf, fini20, kirkpatrick17ewc, aljundi19selfless}, expansion~\cite{rusu16, yoon18, lee20iclr}, and replay~\cite{lopez17, chaudhry19iclr, chaudhry19arxiv, riemer19iclr, javed19}.
Replay-based approaches maintain a fixed-sized memory to rehearse back to the model to mitigate forgetting.
Several works~\cite{lopez17, chaudhry19iclr, chaudhry19arxiv} reserve the space for data samples of previous tasks, while others \cite{shin17} uses a generative model.
Some works \cite{riemer19iclr, javed19} combine rehearsal with meta-learning to find the balance between transfer and interference.
We defer more comprehensive survey including all three branches of continual learning to the appendix.  %

\textbf{Online Sequential Learning}.
In the online sequential learning scenario, a model can only observe the training samples once.
Hence, many works propose methods for maintaining the buffer~\cite{hayes19memory, aljundi19gradient, kim20eccv} or selecting the samples to be rehearsed~\cite{aljundi19mir}.
Recently, \cite{tang21} adopts graphs to represent relational structures between samples, and \cite{gupta20} employs the meta-loss for learning per-parameter learning rates along with model parameters.

Akin to our work, Gdumb~\cite{prabhu19gdumb} and MBPA++~\cite{autume19neurips} also train the model at inference time.
However, greedily selecting samples to be reserved inevitably leads to degradation from noisy labeled data.
Furthermore, discarding the samples that cannot enter the buffer as done in Gdumb may lead to information loss since it only relies on the buffer as its source of training.

\subsection{Noisy Labels}
\label{noisy_related_works}
Learning with noisy labeled data has long been studied~\cite{zhang17under, arpit17, ma18, jiang20}.
Several works design the noise corrected losses~\cite{xu19, harutyunyan20, li19, arazo19, wang19sce} so that the loss minimization of the whole data becomes similar to that of clean samples.
Other works propose to use a noise transition matrix to correct the loss \cite{patrini17, goldberger17, hendrycks18nips, zhang18nips}.
There have been approaches that aim to suppress the contribution of noisy samples by re-weighting the loss~\cite{wang18, ren18l2r}.
Techniques that repair labels~\cite{kremer18a, veit17, li17iccv, song19b, wang19labelrepair, mandal20} or directly learn them~\cite{tanaka18, yi19pencil} are also viable options for learning from noisy labeled data.
Recently, filtering methods based on training dynamics~\cite{huang19, pleiss20aum, nguyen19self} have gained much popularity, based on the observation that models tend to learn clean data first and memorize the noisy labeled data later.  
Small loss sample selection~\cite{jiang17icml, shen19icml, li2020dividemix} techniques by co-teaching \cite{wei20jocor, ge20, han18coteaching, yu19b, malach17, chen19cv} identify noisy samples with multiple models in the same vein.
Some works use graphs for offline learning from a large-scale noisy dataset~\cite{zhang2020global, zhang2021dualgraph}. 
On the other hand, we use a small dataset in the delayed buffer from an online data stream without ground-truth labels; instead we adopt self-supervision to obtain features for the Self-Centered filter.

None of the works mentioned above address continual learning from noisy labeled data streams.
Although \cite{mandal20, li21mopro} also use self-supervised learning with noisy labeled data, they focus on the loss or prediction from the model for selecting suspicious samples.
In the experiments on Table~\ref{tab:exp_filtering}, we will show that training dynamics-based filtering techniques are not viable in noisy labeled continual learning. %
On the other hand, we provide the algorithm that identifies the clean samples while learning from a purified buffer in an online manner.

\subsection{Self-supervised learning}
Self-supervised learning is currently receiving an enormous amount of attention in machine learning research.
The pretext task that trains a model by predicting hidden information within the data includes patch orderings~\cite{doersch15, noroozi16}, image impainting~\cite{pathak16}, colorization~\cite{ye19}, and rotations~\cite{gidaris18, chen19rotation}, to name a few.
There also have been works that utilize the contrastive loss~\cite{chen20icml, he20cvpr, li20proto}; especially, 
SimCLR \cite{chen20icml} proposes a simplified contrastive learning method, which enables representation learning by pulling the randomly transformed samples from the same image closer while pushing ones apart from other images within the batch.
Recently, this instance-wise contrastive learning is extended to prototypical contrastive learning \cite{li20proto} to encode the semantic structures within the data.

\section{Experiments}
\label{experiments}
In our evaluation, we compare SPR with other state-of-the-art models in the online task-free continual learning scenario with label noise.
We test on three benchmark datasets of MNIST~\cite{lecun98}, CIFAR-10~\cite{krizhevsky09} and CIFAR-100~\cite{krizhevsky09} with symmetric and asymmetric random noise, and one large-scale dataset of WebVision~\cite{li17arxiv} with real-world noise on the Web.
We also empirically analyze Self-Replay and the Self-Centered filter from many aspects.

\begin{table*}[t]
    \centering
    \small
    \begin{tabular}{lccc|cc|ccc|cc|c}
        \toprule
        &\multicolumn{5}{c}{MNIST}         &\multicolumn{5}{c}{CIFAR-10} &\multicolumn{1}{c}{WebVision}\\
        &\multicolumn{3}{c|}{symmetric}    &\multicolumn{2}{c|}{asymmetric} &\multicolumn{3}{c|}{symmetric} &\multicolumn{2}{c|}{asymmetric} &\multicolumn{1}{c} {real noise}\\
        noise rate (\%)  & 20 & 40 & 60 & 20 & 40 & 20 & 40 & 60  & 20 & 40 & unknown \\
        \midrule
        Multitask 0\% noise~\cite{caruaca97} & \multicolumn{5}{c|}{98.6} &\multicolumn{5}{c|}{84.7} &\multicolumn{1}{c}{-}  \\
        Multitask~\cite{caruaca97} & 94.5  & 90.5  & 79.8 & 93.4 & 81.1 & 65.6 & 46.7 & 30.0 & 77.0 & 68.7 & 55.5 \\
        \specialrule{0.1pt}{1pt}{1pt}
        Finetune  & 19.3 & 19.0 & 18.7 & 21.1 & 21.1 & 18.5 & 18.1 & 17.0 & 15.3 & 12.4 & 11.9 \\
        \specialrule{0.1pt}{1pt}{1pt}
        EWC~\cite{kirkpatrick17ewc}  & 19.2 & 19.2 & 19.0 & 21.6 & 21.1 & 18.4 & 17.9 & 15.7 & 13.9 & 11.0 & 10.0 \\
        \specialrule{0.1pt}{1pt}{1pt}
        CRS~\cite{vitter85} & 58.6 & 41.8 & 27.2 & 72.3 & 64.2 & 19.6 & 18.5 & 16.8 & 28.9 & 25.2 & 19.3  \\
        CRS + L2R~\cite{ren18l2r} & 80.6 & 72.9 & 60.3 & 83.8 & 77.5 &  29.3 & 22.7 & 16.5 & 39.2 & 35.2 & - \\
        CRS + Pencil~\cite{yi19pencil} & 67.4 & 46.0 & 23.6 & 72.4 & 66.6 & 23.0 & 19.3 & 17.5 & 36.2 & 29.7  & 26.6 \\
        CRS + SL~\cite{wang19sce} & 69.0 & 54.0 & 30.9 & 72.4 & 64.7 & 20.0 & 18.8 & 17.5 & 32.4 & 26.4 & 21.5  \\
        CRS + JoCoR~\cite{wei20jocor} & 58.9 & 42.1 & 30.2 & 73.0 & 63.2 & 19.4 & 18.6 & 21.1 & 30.2 & 25.1 & 19.5  \\

        \specialrule{0.1pt}{1pt}{1pt}
        PRS~\cite{kim20eccv} & 55.5 & 40.2 & 28.5 & 71.5 & 65.6 & 19.1 & 18.5 & 16.7 & 25.6 & 21.6 & 19.0 \\
        PRS + L2R~\cite{ren18l2r} & 79.4 & 67.2 & 52.8 & 82.0 & 77.8 & 30.1 & 21.9 & 16.2 & 35.9 & 32.6 & -  \\
        PRS + Pencil~\cite{yi19pencil} & 62.2 & 33.2 & 21.0 & 68.6 & 61.9 & 19.8 & 18.3 & 17.6 & 29.0 & 26.7 & 26.5   \\
        PRS + SL~\cite{wang19sce} & 66.7 & 45.9 & 29.8 & 73.4 & 63.3 & 20.1 & 18.8 & 17.0 & 29.6 & 24.0 & 21.7  \\
        PRS + JoCoR~\cite{wei20jocor} & 56.0 & 38.5 & 27.2 & 72.7 & 65.5 & 19.9 & 18.6 & 16.9 & 28.4 & 21.9 & 20.2 \\

        \specialrule{0.1pt}{1pt}{1pt}
        MIR~\cite{aljundi19mir}  & 57.9 & 45.6 & 30.9 & 73.1 & 65.7 & 19.6 & 18.6 & 16.4 & 26.4 & 22.1 & 17.2  \\
        MIR + L2R~\cite{ren18l2r} & 78.1 & 69.7 & 49.3 & 79.4 & 73.4 & 28.2 & 20.0 & 15.6 & 35.1 & 34.2 & -  \\
        MIR + Pencil~\cite{yi19pencil} & 70.7 & 34.3 & 19.8 & 79.0 & 58.6 & 22.9 & 20.4 & 17.7 & 35.0 & 30.8 & 22.3  \\
        MIR + SL~\cite{wang19sce} & 67.3 & 55.5 & 38.5 & 74.3 & 66.5 & 20.7 & 19.0 & 16.8& 28.1 & 22.9 & 20.6  \\
        MIR + JoCoR~\cite{wei20jocor} & 60.5 & 45.0 & 32.8 & 72.6 & 64.2 & 19.6 & 18.4 & 17.0 & 27.6 & 23.5 & 19.0  \\

        \specialrule{0.1pt}{1pt}{1pt}
        GDumb~\cite{prabhu19gdumb}  & 70.0 & 51.5 & 36.0 & 78.3 & 71.7 & 29.2 & 22.0 & 16.2 & 33.0 & 32.5 & 30.4  \\
        GDumb + L2R~\cite{ren18l2r} & 65.2 & 57.7 & 42.3 & 67.0 & 62.3 & 28.2 & 25.5 & 18.8 & 30.5 & 30.4 & - \\
        GDumb + Pencil~\cite{yi19pencil} & 68.3 & 51.6 & 36.7 & 78.2 & 70.0 & 26.9 & 22.3 & 16.5 & 32.5 & 29.7 & 26.9  \\
        GDumb + SL~\cite{wang19sce} & 66.7 & 48.6 & 27.7 & 73.4 & 68.1 & 28.1 & 21.4 & 16.3 & 32.7 & 31.8 & 30.8  \\
        GDumb + JoCoR~\cite{wei20jocor} & 70.1 & 56.9 & 37.4 & 77.8 & 70.8 & 26.3 & 20.9 & 15.0 & 33.1 & 32.2 & 24.2  \\
        \specialrule{0.7pt}{1pt}{1pt}

        Self-Centered filter & 80.1 & 79.0 & 77.4 & 80.0    & 79.6   & 36.5     & 35.7 & 32.5 & 37.1 & 36.9 & 33.0 \\
        Self-Replay          & 81.5 & 69.2  & 43.0  & 86.3  & 78.9  & 40.1 & 31.4   & 22.4  & 44.1 & 43.2 & \textbf{48.0} \\
        SPR            & \textbf{85.4}          & \textbf{86.7}          & \textbf{84.8}          & \textbf{86.8}                  & \textbf{86.0}            & \textbf{43.9}          & \textbf{43.0}          & \textbf{40.0}          & \textbf{44.5}          & \textbf{43.9}         & 40.0 \\

        \bottomrule
    \end{tabular}

    \vskip 0.1in
    \caption{\textbf{Overall accuracy} of noisy labeled continual learning after all sequences of tasks are trained. The buffer size is set to 300, 500, 1000 for MNIST, CIFAR-10 and WebVision, respectively.
        Some empty slots on WebVision are due to the unavailability of clean samples required by L2R for training~\cite{ren18l2r}.
        The results are the mean of five unique random seed experiments. We report best performing baselines on different episodes with variances in the appendix.}
    \label{tab:exp_accuracy}
\end{table*}

\subsection{Experimental Design}
\label{experimental_design}
We explicitly ground our experiment setting based on the recent suggestions for robust evaluation in continual learing \cite{aljundi19thesis,farquhar19,vandeven19} as follows.
(i) \textit{Cross-task resemblance}: Consecutive tasks in MNIST~\cite{lecun98}, CIFAR-10~\cite{krizhevsky09}, CIFAR-100~\cite{krizhevsky09}, WebVision~\cite{li17arxiv} are partly correlated to contain neighboring domain concepts. 
(ii) \textit{Shared output heads}: A single output vector is used for all tasks. 
(iii) \textit{No test-time task labels}: Our approach does not require explicit task labels during both training and test phase, often coined as \textit{task-free continual learning} in \cite{aljundi19gradient,lee20iclr,kim20eccv}.
(iv) \textit{More than two tasks}: MNIST~\cite{lecun98}, CIFAR-10~\cite{krizhevsky09}, CIFAR-100~\cite{krizhevsky09} and WebVision~\cite{li17arxiv} contain five, five, twenty, and seven tasks, respectively.

We create a synthetic noisy labeled dataset from MNIST and CIFAR-10 using two methods. %
First, the \textit{symmetric} label noise assigns \{20\%, 40\%, 60\%\} samples of the dataset to other labels within the dataset by a uniform probability. We then create five tasks by selecting random class pairs without replacement.
Second, the \textit{asymmetric} label noise attempts to mimic the real-world label noise by assigning other similar class labels (\eg 5 $\leftrightarrow$ 6, cat $\leftrightarrow$ dog).
We use the similar classes chosen in \cite{patrini17} to contaminate \{20\%, 40\%\} samples of the dataset with similar class pairs. Each task consists of the samples from each corrupted class pair. %
CIFAR-100 has 20 tasks where the \textit{random symmetric} setting has 5 random classes per task with uniform noise across 100 classes. The \textit{superclass symmetric} setting uses
each superclass~\cite{krizhevsky09, lee20iclr} containing 5 classes as a task where the noise is randomized only within the classes in the superclass.
In WebVision, we use the top 14 largest classes in terms of the data size, resulting in 47,784 images in total. We curate seven tasks with randomly paired classes.

We fix the delayed buffer and the replay (purified) buffer size to 300, 500, 1000, 5000 for MNIST, CIFAR-10, WebVision, and CIFAR-100, respectively.
The purified buffer maintains balanced classes as in ~\cite{kim20eccv, prabhu19gdumb}.
We fix the stochastic ensemble size, $E_{max}=5$ unless stated otherwise.
For the base model, we use an MLP with two hidden layers for all MNIST experiments and ResNet-18 for CIFAR-10, CIFAR-100, and WebVision experiments.
Please refer to the appendix for experiment details.

\subsection{Baselines}
\label{baselines}
Since we opt for continual learning from noisy labeled data streams,
we design the baselines combining existing state-of-the-art methods from the two domains of continual learning and noisy label learning.

We explore the replay-based approaches that can learn in the online task-free setting.
We thus choose (i) Conventional Reservoir Sampling (CRS)~\cite{riemer19iclr}, (ii) Maximally Interfered Retrieval (MIR)~\cite{aljundi19mir}, (iii) Partitioning Reservoir Sampling (PRS)~\cite{kim20eccv} and  (iv) GDumb~\cite{prabhu19gdumb}.

For noisy label learning,
we select six models to cover many branches of noisy labeled classification. They include (i) SL loss correction ~\cite{wang19sce}, (ii) semi-supervised JoCoR~\cite{wei20jocor}, (iii) sample reweighting L2R~\cite{ren18l2r}, (iv) label repairing Pencil~\cite{yi19pencil}, (v) training dynamic based detection AUM~\cite{pleiss20aum} and (vi) cross-validation based INCV~\cite{chen19cv}.

\begin{table}[h]
    \centering
    \footnotesize
    \setlength{\tabcolsep}{5.5pt}
    \begin{tabular}{lccc|ccc}
        \toprule
        &\multicolumn{3}{c|}{random symmetric} &\multicolumn{3}{c}{superclass symmetric} \\
        noise rate (\%)  & 20 & 40 & 60 & 20 & 40 & 60 \\
        \midrule
        GDumb + L2R~\cite{ren18l2r} & 15.7 & 11.3 & 9.1 & 16.3 & 12.1 & 10.9 \\
        GDumb + Pencil~\cite{yi19pencil} & 16.7 & 12.5 & 4.1 & 17.5 & 11.6 & 6.8 \\
        GDumb + SL~\cite{wang19sce} & 19.3 & 13.8 & 8.8 & 18.6 & 13.9 & 9.4 \\
        GDumb + JoCoR~\cite{wei20jocor}  & 16.1 & 8.9 & 6.1 & 15.0 & 9.5 & 5.9 \\
        \specialrule{0.4pt}{1pt}{1pt}
        SPR & \textbf{21.5} & \textbf{21.1} & \textbf{18.1} & \textbf{20.5} & \textbf{19.8} & \textbf{16.5}  \\
        \bottomrule
    \end{tabular}
    \vskip 0.1in
    \caption{\textbf{CIFAR100 results} of noisy labeled continual learning after all sequences of tasks are trained.
        The results are the mean of five unique random seed experiments.}
    \label{tab:exp_cifar100}
\end{table}

\begin{table}[h]
    \centering
    \footnotesize
    \setlength{\tabcolsep}{2pt}
    \begin{tabular}{lccc|cc|ccc|cc}
        \toprule
        &\multicolumn{5}{c}{MNIST}  &\multicolumn{5}{c}{CIFAR-10} \\
        &\multicolumn{3}{c|}{symmetric} &\multicolumn{2}{c|}{asymmetric} &\multicolumn{3}{c|}{symmetric} &\multicolumn{2}{c}{asymmetric} \\
        noise rate (\%)  & 20 & 40 & 60 & 20 & 40 & 20 & 40 & 60  & 20 & 40 \\
        \midrule
        AUM~\cite{pleiss20aum} & 7.0 & 16.0 & 11.7 & 30.0 & 29.5 & 36.0 & 24.0 & 11.7 & 46.0 & 30.0 \\
        \specialrule{0.1pt}{1pt}{1pt}
        INCV~\cite{chen19cv} & 23.0 & 22.5 & 14.3 & 37.0 & 31.5 & 22.0 & 18.5 & 9.3 & 37.0 & 30.0  \\
        \specialrule{0.4pt}{1pt}{1pt}
        Non-stochastic & 79.5 & 96.3 & 84.5 & 96.0 & 88.5 & 50.5 & 54.5 & 38.0 & 53.0 & 50.5  \\
        SPR  & \textbf{96.0} & \textbf{96.5} & \textbf{93.0} & \textbf{100} & \textbf{96.5} & \textbf{75.5} & \textbf{70.5} & \textbf{54.3} & \textbf{69.0} & \textbf{60.0} \\
        \bottomrule
    \end{tabular}
    \vskip 0.1in
    \caption{\textbf{Filtered noisy label percentage} in the purified buffer (\eg out of 20\% symmetric noise, SPR filters 96\% of noise). We compare SPR with $E_{max}=5$ to two other state-of-the-art label filtering methods.}
    \label{tab:exp_filtering}
\end{table}

\subsection{Results}
\label{sec:results}
\textbf{Overall performance}. Table~\ref{tab:exp_accuracy} compares the noisy labeled continual learning performance (classification accuracy) between our SPR and baselines on MNIST, CIFAR-10 and WebVision. 
Additionally, Table~\ref{tab:exp_cifar100} compares SPR against the best performing baselines on CIFAR-100 with random symmetric noise and superclass symmetric noise.
SPR performs the best in all symmetric and asymmetric noise types with different levels of 20\%, 40\%, and 60\% as well as real noise.
Multitask is an upper-bound trained with an optimal setting with perfectly clean data (\ie the 0\% noise rate) and offline training. %
Finetune is reported as a lower-bound performance since it performs online training with no continual or noisy label learning technique.

Notably, SPR works much better than L2R~\cite{ren18l2r}, which additionally uses 1000 clean samples for training, giving it a substantial advantage over all the other baselines.
SPR also proves to be much more effective than GDumb~\cite{prabhu19gdumb}, which is the most related method to ours, even when combined with different noisy label learning techniques.

Moreover, the addition of state-of-the-art noisy label techniques is not always beneficial.
This may be because existing noisy label techniques usually assume a large dataset, which is required to reliably estimate the training dynamics to mitigate the noise by regularizing, repairing, and or filtering.
However, the online learning setting is limited by a much smaller dataset (\ie in the purified buffer), leading to a difficult training of the noisy label techniques.

\textbf{Ablation Study}.
To study the effectiveness of each component, two variants of our model that only use Self-Replay or the Self-Centered filter is tested.
That is, the Self-Replay variant does not use any cleaning methods (\ie use conventional reservoir sampling to maintain the purified buffer).
The Self-Centered filter variant finetunes a randomly initialized inference network on the purified buffer instead of finetuning it on the duplicate of the base network.
Both variants outperform all the baselines (excluding L2R) in all three datasets, and combining them our model performs the best on MNIST and CIFAR-10 with all noise levels.
However, WebVision is the only dataset where no synergetic effect is shown, leaving Self-Replay alone to perform the best.
This may be because the WebVision contains highly abstract and noisy classes such as ``Spiral" or ``Cinema," making it difficult for Self-Centered filter to sample from correct clusters.
Please refer to the appendix for further detail.

\textbf{Purification Comparison}.
Table~\ref{tab:exp_filtering} compares the purification performance with the state-of-the-art noise detection methods based on the training dynamics, including AUM~\cite{pleiss20aum} and INCV~\cite{chen19cv}. 
We notice that the performance of AUM and INCV dreadfully declines when detecting label noise among only a small set of data, which is inevitable in online task-free setting,
whereas SPR can filter superbly even with a small set of data.
Even a non-stochastic version of our Self-Centered filter performs better than the baselines. Encouragingly, our method is further improved by introducing stochastic ensembles.

\textbf{Additional Experiments}.
The appendix reports more experimental results, including SPR's noise-free performance, CIFAR-100 filtering performance, episode robustness, purified \& delayed buffer size analysis, ablation of stochastic ensemble size, variance analysis, and data efficiency of Self-Replay.

\section{Conclusion}
\label{conclusion}
We presented the Self-Purified Replay (SPR) framework for noisy labeled continual learning.
At the heart of our framework is Self-Replay, which leverages self-supervised learning to mitigate forgetting and erroneous noisy label signals. 
The Self-Centered filter maintains a purified replay buffer via centrality-based stochastic graph ensembles.
Experiments on synthetic and real-world noise showed that our framework can maintain a very pure replay buffer even with highly noisy data streams while significantly outperforming many combinations of noisy label learning and continual learning baselines.  
Our results shed light on using self-supervision to solve the problems of continual learning and noisy labels jointly. 
Specifically, it would be promising to extend SPR to maintain a not only pure but also more diversified purified buffer.

\section{Acknowledgement}
\label{acknowledgement}
We express our gratitude for the helpful comments on the manuscript by Junsoo Ha, Soochan Lee and the anonymous reviewers for their thoughtful suggestions.
This research was supported by the international coorperation program by NRF of Korea (NRF-2018K2A9A2A11080927), 
Basic Science Research Program through the National Research Foundation of Korea (NRF) (2020R1A2B5B03095585),
and Institue of Information \& communications Technology Planning \& Evaluation (IITP) grant (No.2019-0-01082, SW StarLab).
Gunhee Kim is the corresponding author.

{\small
\bibliographystyle{ieee_fullname}
\bibliography{ref}

\begin{thebibliography}{100}\itemsep=-1pt

\bibitem{aljundi19thesis}
R. Aljundi.
\newblock {\em Continual Learning in Neural Networks}.
\newblock PhD thesis, Department of Electrical Engineering, KU Leuven, 2019.

\bibitem{aljundi19mir}
R. Aljundi, L. Caccia, E. Belilovsky, M. Caccia, M. Lin, L. Charlin, and T.
  Tuytelaars.
\newblock Online continual learning with maximally interfered retrieval.
\newblock In {\em NeurIPS}, 2019.

\bibitem{aljundi19selfless}
R. Aljundi, R. Marcus, and T. Tuytelaars.
\newblock Selfless sequential learning.
\newblock In {\em ICLR}, 2019.

\bibitem{arazo19}
E. Arazo, D. Ortego, P. Albert, N.~E. O'Connor, and K. McGuinness.
\newblock Unsupervised label noise modeling and loss correction.
\newblock In {\em ICML}, 2019.

\bibitem{arpit17}
D. Arpit, S. Jastrzebski, N. Ballas, D. Krueger, E. Bengio, M.~S. Kanwal, T.
  Maharaj, A. Fischer, A. Courville, Y. Bengio, and S. Lacoste-Julien.
\newblock A closer look at memorization in deep networks.
\newblock In {\em ICML}, 2017.

\bibitem{bonacich01}
P Bonacich and P Lloyd.
\newblock Eigenvector-like measures of centrality for asymmetric relations.
\newblock {\em Social Networks}, 3:191--201, 2001.

\bibitem{caron20nips}
M. Caron, I. Misra, J. Marial, P. Goyal, P. Bojanowski, and A. Joulin.
\newblock Unsupervised learning of visual features by contrasting cluster
  assignments.
\newblock In {\em NeurIPS}, 2020.

\bibitem{caruaca97}
R. Caruaca.
\newblock Multitask learning.
\newblock {\em Machine Learning}, 28:41--75, 1997.

\bibitem{chaudhry19iclr}
A. Chaudhry, M. Ranzato, M. Rohrbach, and M. Elhoseiny.
\newblock Efficient lifelong learning with a-gem.
\newblock In {\em ICLR}, 2019.

\bibitem{chaudhry19arxiv}
A. Chaudhry, M. Rohrbach, M. Elhoseiny, T. Ajanthan, P.~K. Dokania, P.~H. Torr,
  and M. Ranzato.
\newblock On tiny episodic memories in continual learning.
\newblock {\em arXiv preprint arXiv:1902.10486v4}, 2019.

\bibitem{chen19cv}
P. Chen, B. Liao, G. Chen, and S. Zhang.
\newblock Understanding and utilizing deep neural networks trained with noisy
  labels.
\newblock In {\em ICML}, 2019.

\bibitem{chen20icml}
T. Chen, S. Kornblith, M. Norouzi, and G. Hinton.
\newblock A simple framework for contrastive learning of visual
  representations.
\newblock In {\em ICML}, 2020.

\bibitem{chen19rotation}
T. Chen, X. Zhai, M. Ritter, M. Lucic, and N. Houlsby.
\newblock Self-supervised gans via auxiliary rotation loss.
\newblock In {\em CVPR}, 2019.

\bibitem{autume19neurips}
C. d'Autume, S. Ruder, L. Kong, and D. Yogatama.
\newblock Episodic memory in lifelong language learning.
\newblock In {\em NeurIPS}, 2019.

\bibitem{dempster77EM}
A.~P. Dempster, N.~M. Laird, and D.~B. Rubin.
\newblock Maximum likelihood from incomplete data via the em algorithm.
\newblock {\em Journal of the Royal Statistical Society}, 1:1--38, 1991.

\bibitem{devlin19bert}
J. Devlin, M.W. Chang, K. Lee, and K. Toutanova.
\newblock Bert: Pre-training of deep bidirectional transformers for language
  understanding.
\newblock In {\em NAACL}, 2019.

\bibitem{doersch15}
C. Doersch, A. Gupta, and A.~A. Efros.
\newblock Unsupervised visual representation learning by context prediction.
\newblock In {\em ICCV}, 2016.

\bibitem{farquhar19}
S. Farquhar and Y. Gal.
\newblock Towards robust evaluations of continual learning.
\newblock {\em arXiv preprint arXiv:1805.09733}, 2019.

\bibitem{fini20}
Enrico Fini, Stéphane Lathuilière, Enver Sangineto, Moin Nabi, and Elisa
  Ricci.
\newblock Online continual learning under extremem memory constraints.
\newblock In {\em ECCV}, 2020.

\bibitem{french99}
R. French.
\newblock Catastrophic forgetting in connectionist networks.
\newblock {\em Trends in Cognitive Sciences}, 3(4):128--135, 1999.

\bibitem{ge20}
Y. Ge, D. Chen, and H. Li.
\newblock Mutual mean-teaching: Pseudo label refinery for unsupervised domain
  adaptation on person re-identification.
\newblock In {\em ICLR}, 2020.

\bibitem{gidaris18}
S. Gidaris, P. Singh, and N. Komodakis.
\newblock Unsupervised representation learning by predicting image rotations.
\newblock In {\em ICLR}, 2018.

\bibitem{goldberger17}
J. Goldberger and E. Ben-Reuven.
\newblock Training deep neural-networks using a noise adaptation layer.
\newblock In {\em ICLR}, 2017.

\bibitem{grill20nips}
J.B. Grill, F. Strub, F. Altche, C. Tallec, P.~H. Richemond, E. Buchatskaya, C.
  Doersch, B.~A. Pires, Z.~D. Guo, M.~G. Azar, B. Piot, K. Kavukcuoglu, R.
  Munos, and M. Valko.
\newblock Big self-supervised models are strong semi-supervised learners.
\newblock In {\em NeurIPS}, 2020.

\bibitem{gupta20}
G. Gupta, K. Yadav, and L. Paull.
\newblock La-maml: Look-ahead meta learning for continual learning.
\newblock In {\em NeurIPS}, 2020.

\bibitem{han18coteaching}
B. Han, Q. Yao, X. Yu, G. Niu, M. Xu, W. Hu, I. Tsang, and M. Sugiyama.
\newblock Co-teaching: Robust training of deep neural networks with extremely
  noisy labels.
\newblock In {\em NeurIPS}, 2018.

\bibitem{wang19labelrepair}
J. Han, P. Luo, and X. Wang.
\newblock Deep self-learning from noisy labels.
\newblock In {\em ICCV}, 2019.

\bibitem{harutyunyan20}
H. Harutyunyan, K. Reing, G.~V. Steeg, and A. Galstyan.
\newblock Improving generalization by controlling label-noise information in
  neural network weights.
\newblock In {\em ICML}, 2020.

\bibitem{hayes19memory}
Tyler~L Hayes, Nathan~D Cahill, and Christopher Kanan.
\newblock Memory efficient experience replay for streaming learning.
\newblock In {\em 2019 International Conference on Robotics and Automation
  (ICRA)}, 2019.

\bibitem{he20cvpr}
K. He, H. Fan, Y. Wu, S. Xie, and R. Girshick.
\newblock Momentum contrast for unsupervised visual representation learning.
\newblock In {\em CVPR}, 2020.

\bibitem{hendrycks18nips}
D. Hendrycks, M. Mazeika, D. Wilson, and K. Gimpel.
\newblock Using trusted data to train deep networks on labels corrupted by
  severe noise.
\newblock In {\em NIPS}, 2018.

\bibitem{huang19}
J. Huang, L. Qu, R. Jia, and B. Zhao.
\newblock O2u-net: A simple noisy label detection approach for deep neural
  networks.
\newblock In {\em ICCV}, 2019.

\bibitem{jacobs1991MoE}
R.~A. Jacobs, M.~I. Jordan, S.~J. Nowlan, and G.~E. Hinton.
\newblock Adaptive mixtures of local experts.
\newblock {\em Neural Comput}, 3:79--87, 1991.

\bibitem{javed19}
K. Javed and M. White.
\newblock Meta-learning representations for continual learning.
\newblock In {\em NeurIPS}, 2019.

\bibitem{jiang20}
L. Jiang, D. Huang, M. Liu, and W. Yang.
\newblock Beyond synthetic noise: Deep learning on controlled noisy labels.
\newblock In {\em ICML}, 2020.

\bibitem{jiang17icml}
L. jiang, Z. Zhou, T. Leung, L. Li, and L. Fei-Fei.
\newblock Mentornet:learning data-driven curriculum for very deep neural
  networks on corrupted labels.
\newblock In {\em ICML}, 2018.

\bibitem{kim20eccv}
D. Kim, J. Jeong, and G. Kim.
\newblock Imbalanced continual learning with partioning reservoir sampling.
\newblock In {\em ECCV}, 2020.

\bibitem{kingma15adam}
D. Kingma and J. Ba.
\newblock {Adam: A Method for Stochastic Optimization}.
\newblock In {\em ICLR}, 2015.

\bibitem{kirkpatrick17ewc}
J. Kirkpatrick, R. Pascanu, N. Rabinowitz, J. Veness, G. Desjardins, A.~A.
  Rusu, K. Milan, J. Quan, T. Ramalho, A. Grabska-Barwinska, D. Hassabis, C.
  Clopath, D. Kumaran, and R. Hadsell.
\newblock Overcoming catastrophic forgetting in neural networks.
\newblock In {\em Proceedings of the National Academy of Sciences}, 2017.

\bibitem{kremer18a}
J. Kremer, F. Sha, and C. Igel.
\newblock Robust active label correction.
\newblock In {\em AISTATS}, 2018.

\bibitem{krizhevsky09}
A. Krizhevsky and G. Hinton.
\newblock Learning multiple layers of features from tiny images.
\newblock Technical report, Computer Science Department, University of Toronto,
  2009.

\bibitem{lecun98}
Y. LeCun, L. Bottou, Y. Bengio, and P. Haffner.
\newblock Gradient based learning applied to document recognition.
\newblock In {\em IEEE}, 1998.

\bibitem{lee18cleannet}
K. Lee, X. He, L. Zhang, and L. Yang.
\newblock Cleannet: Transfer learning for scalable image classfier training
  with label noise.
\newblock In {\em CVPR}, 2018.

\bibitem{lee20iclr}
S. Lee, J. Ha, D. Zhang, and G. Kim.
\newblock A neural dirichlet process mixture model for task-free continual
  learning.
\newblock In {\em ICLR}, 2020.

\bibitem{lesort19b}
T. Lesort, A. Gepperth, A. Stoian, and D. Filliat.
\newblock Marginal replay vs conditional replay for continual learning.
\newblock In {\em IJCANN}, 2019.

\bibitem{li2020dividemix}
Junnan Li, Richard Socher, and Steven~CH Hoi.
\newblock Dividemix: Learning with noisy labels as semi-supervised learning.
\newblock In {\em ICLR}, 2020.

\bibitem{li19}
J. Li, Y. Wong, Q. Zhao, and M. Kankanhalli.
\newblock Learning to learn from noisy labeled data.
\newblock In {\em CVPR}, 2019.

\bibitem{li21mopro}
J. Li, C. Xiong, and S. Hoi.
\newblock Mopro: Webly supervised learning with momentum prototypes.
\newblock In {\em ICLR}, 2021.

\bibitem{li20proto}
J. Li, P. Zhou, C. Xiong, R. Socher, and S.~C.~H. Hoi.
\newblock Prototypical contrastive learning of unsupervised representations.
\newblock In {\em ICLR}, 2020.

\bibitem{li17arxiv}
Wen Li, Limin Wang, Wei Li, Eirikur Agustsson, and Luc~Van Gool.
\newblock Webvision database: Visual learning and understanding from web data.
\newblock {\em arXiv preprint arXiv: 1708.02862}, 2017.

\bibitem{li17iccv}
Y. Li, J. Yang, Y. Song, L. Cao, J. Luo, and L. Li.
\newblock Learning from noisy labels with distillation.
\newblock In {\em ICCV}, 2017.

\bibitem{li16lwf}
Z. Li and D. Hoiem.
\newblock Learning without forgetting.
\newblock In {\em ECCV}, 2016.

\bibitem{lopez17}
D. Lopez-Paz and M. Ranzato.
\newblock Gradient episodic memory for continual learning.
\newblock In {\em NeurIPS}, 2017.

\bibitem{loshchilov17iclr}
I. Loshchilov and F. Hutter.
\newblock Sgdr: Stochastic gradient descent with warm restarts.
\newblock In {\em ICLR}, 2017.

\bibitem{lukasik20}
M. Lukasik, S. Bhojanapalli, A.~K. Menon, and S. Kumar.
\newblock Does label smoothing mitigate label noise?
\newblock In {\em ICML}, 2020.

\bibitem{lyu20iclr}
Y. Lyu and I.W. Tsang.
\newblock Curriculum loss: Robust learning and generalization against label
  corruption.
\newblock In {\em ICLR}, 2020.

\bibitem{ma18}
X. Ma, Y. Wang, M.~E. Houle, S. Zhou, S.~M. Erfani, S. Xia, S. Wijewickrema,
  and J. Bailey.
\newblock Dimensionality-driven learning with noisy labels.
\newblock In {\em ICML}, 2018.

\bibitem{malach17}
E. Malach and S. Shalev-Shwartz.
\newblock Decoupling "when to update" from "how to update".
\newblock In {\em NeurIPS}, 2017.

\bibitem{mandal20}
D. Mandal, S. Bharadwaj, and S. Biswas.
\newblock A novel self-supervised re-labeling approach for training with noisy
  labels.
\newblock In {\em WACV}, 2020.

\bibitem{mccloskey89}
M. McCloskey and N.~J. Cohen.
\newblock Catastrophic interference in conncectionist networks.
\newblock {\em Psychology of learning and motivation}, 24:109--265, 1989.

\bibitem{nguyen19self}
D.~T. Nguyen, C.~K. Mummadi, T.~P.~N. Ngo, T.~H.~P. Nguyen, L. Beggel, and T.
  Brox.
\newblock Self: Learning to filter noisy labels with self-ensembling.
\newblock In {\em ICLR}, 2019.

\bibitem{nieminen74}
J Nieminen.
\newblock On the centrality in a graph.
\newblock {\em Scandinavian Journal of Psychology}, 1:332--336, 1974.

\bibitem{noroozi16}
M. Noroozi and P. Favaro.
\newblock Unsupervised learning of visual representations by solving jigsaw
  puzzles.
\newblock In {\em ECCV}, 2017.

\bibitem{ostyakov18eccv}
P. Ostyakov, E. Logacheva, R. Suvorov, V. Aliev, G. Sterkin, O. Khomenko, and
  S.~I. Nikolenko.
\newblock Label denoising with large ensembles of heterogeneous neural
  networks.
\newblock In {\em ECCV}, 2018.

\bibitem{pathak16}
D. Pathak, P. Krahenbuhl, J. Donahue, T. Darrell, and A.~A. Efros.
\newblock Context encoders: Feature learning by inpainting.
\newblock In {\em CVPR}, 2016.

\bibitem{patrini17}
G. Patrini, A. Rozza, A. Menon, R. Nock, and L. Qu.
\newblock Making deep neural networks robust to label noise: a loss correction
  approach.
\newblock In {\em CVPR}, 2017.

\bibitem{pleiss20aum}
G. Pleiss, T. Zhang, E.~R. Elenberg, and K.~Q. Weinberger.
\newblock Identifying mislabeled data using the area under the margin ranking.
\newblock In {\em NIPS}, 2020.

\bibitem{prabhu19gdumb}
A. Prabhu, P.~H.S. Torr, and P.~K. Dokania.
\newblock Gdumb: A simple approach that questions our progress in continual
  learning.
\newblock In {\em ECCV}, 2019.

\bibitem{aljundi19gradient}
Aljundi Rahaf, Min Lin, Baptiste Goujaud, and Bengio Yoshua.
\newblock Gradient based sample selection for online continual learning.
\newblock In {\em NeurIPS}, 2019.

\bibitem{ratcliff90}
R. Ratcliff.
\newblock Conncectionist models of recognition memory: Constraints imposed by
  learning and forgetting functions.
\newblock {\em Pscyhological review}, 97(2):285--308, 1990.

\bibitem{reed15}
S. Reed, H. Lee, D. Anguelov, C. Szegedy, D. Erhan, and A. Rabinovich.
\newblock Training deep neural networks on noisy labels with bootstrapping.
\newblock In {\em ICLR workshop}, 2015.

\bibitem{ren18l2r}
M. Ren, W. Zeng, B. Yang, and R. Urtasun.
\newblock Learning to reweight examples for robust deep learning.
\newblock In {\em ICML}, 2018.

\bibitem{riemer19iclr}
M. Riemer, I. Cases, R. Ajemian, M. Liu, I. Rish, Y. Tu, and G. Tesauro.
\newblock Learning to learn without forgetting by maximizing transfer and
  minimizing interference.
\newblock In {\em ICLR}, 2019.

\bibitem{rolnick19}
D. Rolnick, A. Ahuja, J. Schwarz, T.~P. Lillicrap, and G. Wayne.
\newblock Experience replay for continual learning.
\newblock In {\em NeurIPS}, 2019.

\bibitem{rusu16}
A.~A. Rusu, N.~C. Rabinowitz, G. Desjardins, H. Soyer, J. Kirkpatrick, K.
  Kavukcuoglu, R. Pascanu, and R. Hadsell.
\newblock Progressive neural networks.
\newblock {\em arXiv preprint arXiv:1606.04671}, 2016.

\bibitem{shen19icml}
Y. Shen and S. Sanghavi.
\newblock Learning with bad training data via iterative trimmed loss
  minimization.
\newblock In {\em ICML}, 2019.

\bibitem{shin17}
H. Shin, J.~K. Lee, J. Kim, and J. Kim.
\newblock Continual learning with deep generative replay.
\newblock In {\em NeurIPS}, 2017.

\bibitem{silva2017cuda}
Gustavo Rodrigues~Lacerda Silva, Rafael~Ribeiro De~Medeiros, Brayan
  Rene~Acevedo Jaimes, Carla~Caldeira Takahashi, Douglas Alexandre~Gomes
  Vieira, and Ant{\^o}Nio De~P{\'a}Dua Braga.
\newblock Cuda-based parallelization of power iteration clustering for large
  datasets.
\newblock {\em IEEE Access}, 5:27263--27271, 2017.

\bibitem{song19b}
H. Song, M. Kim, and J. Lee.
\newblock Selfie: Refurbishing unclean samples for robust deep learning.
\newblock In {\em ICML}, 2019.

\bibitem{squire81}
L.~R. Squire.
\newblock Two forms of human amnesia: an analysis of forgetting.
\newblock {\em Journal of Neuroscience}, 6:635--640, 1981.

\bibitem{tanaka18}
D. Tanaka, D. Ikami, T. Yamasaki, and K. Aizawa.
\newblock Joint optimization framework for learning with noisy labels.
\newblock In {\em CVPR}, 2018.

\bibitem{tang21}
B. Tang and D.~S. Matteson.
\newblock Graph-based continual learning.
\newblock In {\em ICLR}, 2021.

\bibitem{thrun96}
S. Thrun.
\newblock Is learning the n-th thing any easier than learning the first?
\newblock In {\em Advances in neural information processing systems}, 1996.

\bibitem{vandeven19}
G.~M. van~de Ven and S.~T. Andreas.
\newblock Three scenarios for continual learning.
\newblock In {\em NeurIPS Continual Learning workshop}, 2019.

\bibitem{veit17}
A. Veit, N. Alldrin, G. Chechik, I. Krasin, A. Gupta, and S. Belongie.
\newblock Learning from noisy large-scale datasets with minimal supervision.
\newblock In {\em CVPR}, 2017.

\bibitem{vitter85}
J.~S. Vitter.
\newblock Random sampling with a reservoir.
\newblock {\em ACM Transactions on Mathematical Software (TOMS)}, 11(1):37--57,
  1985.

\bibitem{vonMises1929}
R. von Mises and H. Pollaczek-Geiringer.
\newblock Practical methods of solving equations.
\newblock {\em Journal of Applied Mathematics and Mechanics}, 9:152--164, 1929.

\bibitem{wang18}
Y. Wang, W. Liu, X. Ma, J. Bailey, H. Zha, L. Song, and S. Xia.
\newblock Iterative learning with open-set noisy labels.
\newblock In {\em CVPR}, 2018.

\bibitem{wang19sce}
Y. Wang, X. Ma, Z. Chen, Y. Luo, J. Yi, and J. Bailey.
\newblock Symmetric cross entropy for robust learning with noisy labels.
\newblock In {\em ICCV}, 2019.

\bibitem{wei20jocor}
H. Wei, L. Feng, X. Chen, and B. An.
\newblock Combating noisy labels by agreement: A joint training method with
  co-regularization.
\newblock In {\em CVPR}, 2020.

\bibitem{xu19}
Y. Xu, P. Cao, Y. Kong, and Y. Wang.
\newblock L\_dmi: An information-theoretic noise-robust loss function.
\newblock In {\em NeurIPS}, 2019.

\bibitem{ye19}
M. Ye, X. Zhang, P.~C. Yuen, and S. Chang.
\newblock Unsupervised embedding learning via invariant and spreading instance
  feature.
\newblock In {\em CVPR}, 2019.

\bibitem{yi19pencil}
K. Yi and J. Wu.
\newblock Probabilistic end-to-end noise correction for learning with noisy
  labels.
\newblock In {\em CVPR}, 2019.

\bibitem{yoon18}
J. Yoon, E. Yang, J. Lee, and S.~J. Hwang.
\newblock Lifelong learning with dynamically expandable networks.
\newblock In {\em ICLR}, 2018.

\bibitem{yu19b}
X. Yu, B. Han, J. Yao, G. Niu, I. Tsang, and M. Sugiyama.
\newblock How does disagreement help generalization against label corruption?
\newblock In {\em ICML}, 2019.

\bibitem{yun19cutmix}
S. Yun, D. Han, S.~J. Oh, S. Chun, J. Choe, and Y. Yoo.
\newblock Cutmix: Regularization strategy to train strong classiﬁers with
  localizable features.
\newblock In {\em ICCV}, 2019.

\bibitem{zenke17}
F. Zenke, B. Poole, and S. Ganguli.
\newblock Continual learning through syanptic intelligence.
\newblock In {\em ICML}, 2017.

\bibitem{zhang17under}
C. Zhang, S. Bengio, M. Hardt, B. Recht, and O. Vinyals.
\newblock Understanding deep learning requires rethinking generalization.
\newblock In {\em ICLR}, 2017.

\bibitem{zhang2021dualgraph}
HaiYang Zhang, XiMing Xing, and Liang Liu.
\newblock Dualgraph: A graph-based method for reasoning about label noise.
\newblock In {\em CVPR}, 2021.

\bibitem{zhang17er}
S. Zhang and R. Sutton.
\newblock A deeper look at experience replay.
\newblock {\em arXiv preprint arXiv:1712.01275}, 2017.

\bibitem{zhang2020global}
Yaobin Zhang, Weihong Deng, Mei Wang, Jiani Hu, Xian Li, Dongyue Zhao, and
  Dongchao Wen.
\newblock Global-local gcn: Large-scale label noise cleansing for face
  recognition.
\newblock In {\em CVPR}, 2020.

\bibitem{zhang18nips}
Z. Zhang and M. Sabuncu.
\newblock Generalized cross entropy loss for training deep neural networks with
  noisy labels.
\newblock In {\em NIPS}, 2018.

\end{thebibliography}
}
\clearpage

\begin{appendix}

{\Large \bf{Appendix}}\\

\ificcvfinal\thispagestyle{empty}\fi

This appendix enlists the following additional materials. %

\begin{enumerate}[label=\Roman*.]
    \item Posterior in the Beta Mixture Model. Sec.~\ref{sec:bmm_fitting}
    \item Extended Related Work. Sec.~\ref{sec:add_related}
    \item Experiment Details. Sec.~\ref{sec:add_experiments}
    \item Extended Results \& Analyses. Sec.~\ref{sec:add_analysis}
    \begin{enumerate}[label=\roman*.]
        \item Efficiency of Eigenvector Centrality. Sec.~\ref{sec:eff_eig}
        \item Noise-Free Performance. Sec.~\ref{sec:noise_free}
        \item Noise Robustness Comparison. Sec.~\ref{sec:noise_robust}
        \item Features from ImageNet Pretrained Model. Sec.~\ref{sec:imgpre_filter}
        \item Analyses of Stochastic Ensemble Size ($E_{max}$). Sec.~\ref{sec:ensemble}
        \item Filtering performances on CIFAR-100. Sec.~\ref{sec:cifar100_filter}
        \item Self-Replay with Noisy Labeled Data. Sec.~\ref{sec:data_efficiency}
        \item Analyses of the Results on WebVision. Sec.~\ref{sec:ext_ablation_study}
        \item Episode Robustness. Sec.~\ref{sec:episode_robust}
        \item Buffer Size Analysis. Sec.~\ref{sec:buffsize_analysis}
        \item Variance. Sec.~\ref{sec:variance}
    \end{enumerate}

\end{enumerate}

\section{Posterior in the Beta Mixture Model}
\label{sec:bmm_fitting}
We provide some details about how to fit beta mixture models~\cite{jacobs1991MoE} with the EM-algorithm~\cite{dempster77EM} to obtain the posterior $p(z|c)$ for the central point with score $c$.
In the E-step, fixing $\pi_z, \alpha_z, \beta_z$, we update the latent variables using the Bayes rule:
\begin{align}
\gamma_z(c) = p(z|c) =  \frac{\pi_z p(c|\alpha_z, \beta_z)}{\sum_{j=1}^Z \pi_j p(c|\alpha_j, \beta_j)}.
\end{align}
In the M-step, fixing the posterior $\gamma_z(c)$, we estimate the distribution parameters $\alpha$ and $\beta$ using method of moments:
\begin{align}
    \alpha_z = \bar{c}_z (\frac{\bar{c}_z ( 1-\bar{c}_z)}{s^2_z} - 1), \ \ 
    \beta_z = \frac{\alpha_z (1 - \bar{c}_z)}{\bar{c}_z},
\end{align}
where $\bar{c}_z$ is the weighted average of the centrality scores from all the points in the delayed batch, and
$s^2_z$ is the weighted variance estimate as
\begin{align}
\bar{c}_z &= \frac{\sum^N_{i=1} \gamma_z(c_i)c_i}{\sum_{i=1}^N \gamma_z (c_i)}, \\
s^2_z &= \frac{\sum^N_{i=1} \gamma_z(c_i)(c_i - \bar{c}_z)^2}{\sum_{i=1}^N \gamma_z(c_i)}, \\
\pi_z &= \frac{1}{N} \sum_{i=1}^N \gamma_z(c_i).
\end{align}
Finally, we arrive at $p(z|c) \propto p(z)p(c|z)$.

\section{Extended Related Work}
\label{sec:add_related}
\subsection{Continual Learning}
\label{continual_related_works}
Continual learning is mainly tackled from three main branches of regularization, expansion, and replay.

\textbf{Regularization-based Approaches}.
Methods in this branch prevent forgetting by penalizing severe drift of model parameters.
Learning without Forgetting \cite{li16lwf} employs knowledge distillation to preserve the previously learned knowledge.
Similarly, MC-OCL \cite{fini20} proposes batch-level distillation to balance stability and plasticity in an online manner.
Elastic Weight Consolidation \cite{kirkpatrick17ewc} finds the critical parameters for each task by applying the Fisher information matrix. %
Recently, Selfless Sequential Learning \cite{aljundi19selfless} enforces representational sparsity, reserving the space for future tasks.

\textbf{Expansion-based Approaches}.
Many methods in this branch explicitly constrain the learned parameters by freezing the model and instead allocate additional resources to learn new tasks.
Progressive Neural Network \cite{rusu16} prevent forgetting by prohibiting any updates on previously learned parameters while allocating new parameters for the training of the future tasks.
Dynamically Expandable Networks\cite{yoon18} decides on the number of additional neurons for learning new tasks using $L2$ regularization for sparse and selective retraining.
CN-DPM \cite{lee20iclr} adopts the Bayesian nonparametric framework to expand the model in an online manner.

\textbf{Replay-based Approaches}.
The replay-based branch maintains a fixed-sized memory to rehearse back to the model to mitigate forgetting.
The fixed-sized memory could be in the form of a buffer for the data samples of previous tasks or the form of generative model weights \cite{shin17} to generate the previous tasks' data.
GEM~\cite{lopez17} and AGEM~\cite{chaudhry19iclr} use a buffer to constrain the gradients in order to alleviate forgetting.
In \cite{chaudhry19arxiv}, training a model even on tiny episodic memory can achieve an impressive performance.
Some recent approaches\cite{riemer19iclr, javed19} combine rehearsal with meta-learning to find the balance between transfer and interference.

\textbf{Online Sequential Learning}.
Online sequential learning is closely related to continual learning research, as it assumes that a model can only observe the training samples once before discarding them.
Thus, it is a fundamental problem to maintain the buffer or selecting the samples to be rehearsed.
ExStream~\cite{hayes19memory} proposes the buffer maintenance method by clustering the data in an online manner.
GSS~\cite{aljundi19gradient} formulates sample selection for the buffer as a constraint reduction, while MIR~\cite{aljundi19mir} proposes a sample retrieving method from the buffer by selecting the most interfered samples.
Considering real-world data are often imbalanced and multi-labeled, PRS~\cite{kim20eccv} tackles this problem by partitioning the buffer for each class and maintaining it to be balanced.
Also, combining graphs or meta-learning with online continual learning has been studied.
Graphs are adopted to represent the relational structures between samples \cite{tang21}, and the meta-loss is applied for learning not only model weights but also per-parameter learning rates \cite{gupta20}.
Recently, GDumb~\cite{prabhu19gdumb} and MBPA++~\cite{autume19neurips} show training a model at inference time improves the overall performance.

\subsection{Noisy Labels}
\label{noisy_related_works}
Learning with noisy labeled data has been a long-studied problem.
In several works \cite{zhang17under, arpit17, ma18} make an important empirical observation that DNNs usually learn the clean data first then subsequently memorize the noisy data.
Recently, a new benchmark \cite{jiang20} has been proposed to simulate real-world label noise from the Web. 
Noisy labeled data learning can be categorized into loss regularization, data re-weighting, label cleaning, clean sample selection via training dynamics.

\textbf{Loss Regularization}.
This approach designs the noise correction loss so that the optimization objective is equivalent to learning with clean samples.
\cite{patrini17} proposes using a noise transition matrix for loss correction. \cite{goldberger17} appends a new layer to DNNs to estimate the noise transition matrix while \cite{hendrycks18nips} additionally uses a small set of clean data. \cite{zhang18nips} studies a set of theoretically grounded noise-robust loss functions that can be considered a generalization of the mean absolute error and categorical cross-entropy. \cite{xu19, harutyunyan20} propose new losses based on information theory. \cite{li19} adopts the meta-loss to find noise-robust parameters. \cite{arazo19} uses a bootstrapping loss based on the estimated noise distribution.

\textbf{Data Re-weighting}.
This approach suppresses the contribution of noisy samples by re-weighting the loss.
\cite{ren18l2r} utilizes meta-learning to estimate example importance with the help of a small clean data.
\cite{wang18} uses a Siamese network to estimate sample importance in an open-set noisy setting.

\textbf{Label Cleaning}.
This approach aims at explicitly repairing the labels.
\cite{lukasik20} shows that using smooth labels is beneficial in noisy labeled data learning. \cite{tanaka18, yi19pencil} propose to learn the data labels as well as the model parameters.
\cite{reed15, song19b} re-label the samples using the model predictions.
Additionally, \cite{kremer18a} adopts the active learning strategy to choose the samples to be re-labeled.
\cite{veit17, li17iccv, ge20} employ multiple models, while \cite{wang19labelrepair, mandal20, li21mopro} utilize prototypes to refine the noisy labels.

\textbf{Training Procedures}.
Following the observations that clean data and easy patterns are learned prior to noisy data~\cite{zhang17under, arpit17, ma18}, several works propose filtering methods based on model training dynamics.
\cite{jiang17icml} adopts curriculum learning by selecting small loss samples.
\cite{wei20jocor, ge20, han18coteaching, yu19b, malach17} identify the clean samples using losses or predictions from multiple models and feed them into another model.
\cite{huang19, pleiss20aum, nguyen19self} filter noisy samples based on the accumulated losses or predictions.
\cite{chen19cv} proposes to fold the training data and filter clean samples by cross-validating those split data.

\subsection{Self-supervised Learning}
Self-supervised learning enables the training of a model to utilize its own unlabeled inputs and often shows remarkable performance on downstream tasks.
One example of self-supervised learning uses a pretext task, which trains a model by predicting the data's hidden information.
Some examples include patch orderings~\cite{doersch15, noroozi16}, image impainting~\cite{pathak16}, colorization~\cite{ye19}, and rotations~\cite{gidaris18, chen19rotation}.
Besides designing heuristic tasks for self-supervised learning, some additional works utilize the contrastive loss.
\cite{chen20icml} proposes a simpler contrastive learning method, which performs representation learning by pulling the randomly transformed samples closer while pushing them apart from the other samples within the batch.
\cite{he20cvpr} formulates contrastive learning as a dictionary look-up and uses the momentum-updated encoder to build a large dictionary.
Recently, \cite{li20proto} extends instance-wise contrastive learning to prototypical contrastive learning to encode the semantic structures within the data.

\section{Experiment Details}
\label{sec:add_experiments}

We present the detailed hyperparameter setting of SPR training as well as the baselines.
We resize the images into $28 \times 28$ for MNIST~\cite{lecun98}, $32 \times 32$ for CIFAR-10~\cite{krizhevsky09}, and $84 \times 84$ for WebVision~\cite{li17arxiv}.
We set the size of delayed and purified buffer to 300 for MNIST, 500 for CIFAR-10, and 1000 for WebVision on all methods.
We use the batch size of self-supervised learning as 300 for MNIST, 500 for CIFAR-10, and 1000 on WebVision. The batch size of supervised learning is fixed to 16 for all experiments. 
The number of training epochs for the base and expert network are respectively 3000 and 4000 on all datasets, while finetuning epochs for the inference network is 50.
The NTXent loss~\cite{chen20icml} uses a temperature of $0.5$, and $E_{max}=5$ for SPR.
We use the Adam optimizer~\cite{kingma15adam} with setting $\beta_1 = 0.9$, $\beta_2=0.999$, $\epsilon=0.0002$ for self-supervised training of both base and expert network, and $\epsilon=0.002$ for supervised finetuning.

The hyperparameters for the baselines are as follows. 
\begin{enumerate}[label=$\bullet$]%
\item Multitask~\cite{caruaca97}: We perform i.i.d offline training for 50 epochs with uniformly sampled mini-batches.
\item Finetune: We run online training through the sequence of tasks.
\item GDumb~\cite{prabhu19gdumb}: As an advantage to GDumb, we allow CutMix~\cite{yun19cutmix} with $p=0.5$ and $\alpha=1.0$. We use the SGDR~\cite{loshchilov17iclr} schedule with $T_0=1$ and $T_{mult}=2$. Since access to a validation data in task-free continual learning is not natural, the number of epochs is set to 100 for MNIST and CIFAR-10 and 500 for WebVision.
\item PRS~\cite{kim20eccv}: We set $\rho = 0$.
\item L2R~\cite{ren18l2r}: We use meta update with $\alpha=1$, and set the number of clean data per class as 100 and the clean update batch size as 100.
\item Pencil~\cite{yi19pencil}: We use $\alpha=0.4$, $\beta=0.1$, stage1 = 70, stage2 = 200, $\lambda=600$.
\item SL~\cite{wang19sce}: We use $\alpha=1.0$, $\beta=1.0$.
\item JoCoR~\cite{wei20jocor}: We set $\lambda=0.1$.
\item AUM~\cite{pleiss20aum}: We set the learning rate to 0.1, momentum to 0.9, weight decay to 0.0001 with  a batch size of 64 for 150 epochs. We apply random crop and random horizontal flip for input augmentation.
\item INCV~\cite{chen19cv}: We set the learning rate to 0.001, weight decay to 0.0001, a batch size 128 with 4 iterations for 200 epochs. We apply random crop and random horizontal flip for input augmentation.
\end{enumerate}

\section{Extended Results \& Analyses}
\label{sec:add_analysis}
We provide more in-depth results and analyses of the experiments in this section.
\subsection{Efficiency of Eigenvector Centrality}
\label{sec:eff_eig}
The time and space complexity of Eigenvector centrality is $O(n^2)$, where $n$ is the number of data.
Our online scenario constraints the size of $n$ (Delayed buffer size) to be less than $2\%$ of the entire dataset.
Also, for $k$ classes, the complexity reduces to $O((n/k)^2 )$ since the Self-Centered filter computes per class.
On Quadro RTX GPU, building the adjacency matrices took less than $0.0003s$.
On a CPU, Eigenvector centrality computation took $0.4s, 1.3s, 7.1s$ for buffers of $300, 500, 1K$, respectively, which can speed up to $188×$ by GPU~\cite{silva2017cuda}.

\subsection{Noise-Free Performance}
\label{sec:noise_free}
Table~\ref{tab:noise_free} compares our SPR and Self-Replay's performance against Gdumb's reported performances on MNIST and CIFAR-10.
Interestingly, our Self-Replay performs better than Gdumb, showing great promise in the direction of self-superved continual learning in general.
However, SPR's performance is below that of Gdumb when completely noise free.
We speculate SPR's mechanics to retain clean samples lead to a tradeoff with precise class feature coverage which seems to be of relative importance in a noise-free setting.

\begin{table}[h]
    \centering
    \small
    \setlength{\tabcolsep}{5pt}
    \begin{tabular}{lcc}
        \toprule
        & MNIST &CIFAR-10 \\
        \midrule
        Gdumb~\cite{prabhu19gdumb} & \textbf{91.9} & 45.8 \\
        \specialrule{0.1pt}{1pt}{1pt}
        Self-Replay & 88.9 & \textbf{47.4} \\
        SPR & 85.5 & 44.3 \\
        \bottomrule
    \end{tabular}
    \vskip 0.1in
    \caption{\textbf{Noise Free performances} of Self-Replay and SPR compared with Gdumb~\cite{prabhu19gdumb}'s reported performances.
    Buffer size is fixed to 500.}
    \label{tab:noise_free}
\end{table}

\subsection{Noise Robustness Comparison}
\label{sec:noise_robust}
\begin{figure}[t]
\begin{center}
\includegraphics[width=\columnwidth]{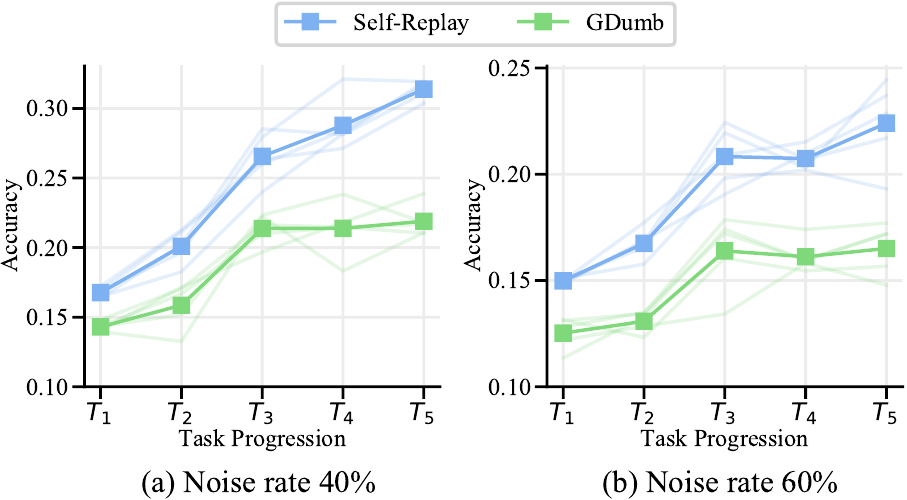}
\caption{\textbf{Noise Robustness} of Self-Replay and GDumb on CIFAR-10. Both models use conventional reservoir sampling (\ie uniform random sampling from the input data stream) for the replay (purified) buffer;  that is, no purification of the input data is performed. The vivid plots indicate the mean of five random seed experiments.}
\label{fig:SRvsGD}
\end{center}
\end{figure}

Figure~\ref{fig:SRvsGD} contrasts the noise robustness of the strongest and closest baseline GDumb to Self-Replay under 40\% and 60\% noise levels while removing the Self-Centered filter from our method.  %
Even still, Self-Replay is much more robust against high amounts of noisy labels at every task, validating that Self-Replay alone is able to mitigate the harmful effects of noise to a great extent.

\subsection{Features from ImageNet Pretrained Model}
\label{sec:imgpre_filter}
We would like to clarify that our scenario and approach is much different and novel in that, 
the algorithm assumes an online stream of data and no ground-truth data is available to supervisedly train a noise detector. 
Not only that, the data we have to work is very small (e.g., $300, 500, 1000$) as the purpose is for a Delayed buffer to set aside small amounts from a stream of data for verification by our self-supervisedly trained Expert model. 
This was also motivated by the empirical evidence that using a supervised learning technique such as AUM~\cite{pleiss20aum}, INCV~\cite{chen19cv}, and using an ImageNet supervisedly pre-trained model for extracting the features led to worthless performances in the Table~\ref{tab:exp_filtering_imgpre}, \ref{tab:exp_filtering}.
\begin{table}[h]
    \centering
    \small
    \setlength{\tabcolsep}{5pt}
    \begin{tabular}{lccc}
        \toprule
        &\multicolumn{3}{c}{CIFAR-10} \\
        &\multicolumn{3}{c}{symmetric} \\
        noise rate (\%)  & 20 & 40 & 60 \\
        \midrule
        ImageNet pretrained & -9.0 & -7.0 & 3.0 \\
        \specialrule{0.1pt}{1pt}{1pt}
        Self-supervised & 75.5 & 70.5 & 54.3 \\
        \bottomrule
    \end{tabular}
    \vskip 0.1in
    \caption{\textbf{Filtered noisy label percentages} in the purified buffer. 
    We compare filtering performances from the self-supervisedly learned features with the ones from the ImageNet pretrained features. We set $E_{max}=5$.}
    \label{tab:exp_filtering_imgpre}
\end{table}

\subsection{Analyses of Stochastic Ensemble Size ($E_{max}$)}
\label{sec:ensemble}
Figure~\ref{fig:stochastic_ensemble_illust} displays the performance of Stochastic Ensemble by increasing the ensemble sizes ($E_{max}$) from 1 to 40.
Stochastic Ensemble performs better in all ensemble sizes than the non-stochastic BMM in terms of the percentages of filtered noisy labels on both MNIST and CIFAR10 with 60\% noisy labels.
A substantial boost is seen in the filtering performance up to 10. After 20, the performance starts to plateau on both MNIST and CIFAR-10.
The empirically suggested optimal number of $E_{max}$ may be around 20 for both MNIST and CIFAR-10 and this is further confirmed in Table~\ref{tab:exp_filtering} where we fix $E_{max}=20$ and the overall filtering percentage increase by 2.4\% on average, compared to the results in the main draft with $E_{max} = 5$.

\begin{figure}[h]
\begin{center}
    \includegraphics[width=\columnwidth]{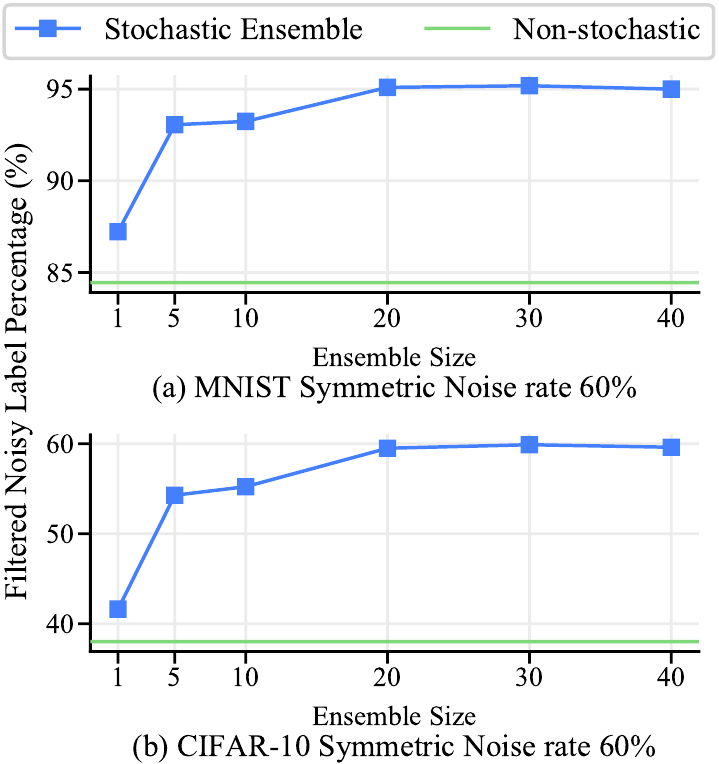}
\end{center}
\caption{\textbf{Filtered noisy label percentages} in the purified buffer by increasing the ensemble size ($E_{max}$) on MNIST and CIFAR-10 with 60\% noise rate. Stochastic Ensemble significantly performs better than the static version. }
\label{fig:stochastic_ensemble_illust}
\end{figure}

\begin{table}[t]
    \centering
    \footnotesize
    \setlength{\tabcolsep}{2pt}
    \begin{tabular}{lccc|cc|ccc|cc}
        \toprule
        &\multicolumn{5}{c}{MNIST}  &\multicolumn{5}{c}{CIFAR-10} \\
        &\multicolumn{3}{c|}{symmetric} &\multicolumn{2}{c|}{asymmetric} &\multicolumn{3}{c|}{symmetric} &\multicolumn{2}{c}{asymmetric} \\
        noise rate (\%)  & 20 & 40 & 60 & 20 & 40 & 20 & 40 & 60  & 20 & 40 \\
        \midrule
        AUM~\cite{pleiss20aum} & 7.0 & 16.0 & 11.7 & 30.0 & 29.5 & 36.0 & 24.0 & 11.7 & 46.0 & 30.0 \\
        \specialrule{0.1pt}{1pt}{1pt}
        INCV~\cite{chen19cv} & 23.0 & 22.5 & 14.3 & 37.0 & 31.5 & 22.0 & 18.5 & 9.3 & 37.0 & 30.0  \\
        \specialrule{0.4pt}{1pt}{1pt}
        Non-stochastic & 79.5 & 96.3 & 84.5 & 96.0 & 88.5 & 50.5 & 54.5 & 38.0 & 53.0 & 50.5  \\
        SPR (Ours) & \textbf{95.0} & \textbf{96.8} & \textbf{95.0} & \textbf{99.9} & \textbf{97.5} & \textbf{79.5} & \textbf{76.3} & \textbf{59.5} & \textbf{72.0} & \textbf{59.0} \\

        \bottomrule
    \end{tabular}
    \vskip 0.1in
    \caption{\textbf{Filtered noisy label percentages} in the purified buffer. We compare SPR to two other state-of-the-art label filtering methods. We set $E_{max}=20$.}
    \label{tab:exp_filtering}
\end{table}

\subsection{Filtering performances on CIFAR-100.}
\label{sec:cifar100_filter}
Table~\ref{tab:exp_cifar100_filter} compares the filtering performances of SPR with the two state-of-the-art label filtering methods~\cite{pleiss20aum, chen19cv} on CIFAR-100.
SPR performs the best in all random symmetric noise and superclass symmetric noise with different levels of 20\%, 40\%, and 60\%.
Even the filtering performance on CIFAR-100 is superior to CIFAR-10. We believe this result is mainly due to the classes in CIFAR100 being more specific than CIFAR10
(e.g., automobile, airplane, bird in CIFAR10 where CIFAR100 has the trees superclass divided into maple, oak, palm, pine, willow), allowing SPR to self-supervisedly learn much more distinct features per class.
This result is further reinforced on the WebVision dataset where SPR shows a weakness in filtering abstract classes such as ``Spiral,” in which the details can be found in Sec~\ref{sec:ext_ablation_study}.

\begin{table}[h]
    \centering
    \small
    \setlength{\tabcolsep}{5.5pt}
    \begin{tabular}{lccc|ccc}
        \toprule
        &\multicolumn{3}{c|}{random symmetric} &\multicolumn{3}{c}{superclass symmetric} \\
        noise rate (\%)  & 20 & 40 & 60 & 20 & 40 & 60 \\
        \midrule
        AUM~\cite{pleiss20aum} & 33.5 & 46.8 & 13.9 & 25.0 & 21.4 & 32.4 \\
        \specialrule{0.1pt}{1pt}{1pt}
        INCV~\cite{chen19cv} & 46.9 & 34.8 & 22.2 & 33.7 & 27.0 & 15.4 \\
        \specialrule{0.4pt}{1pt}{1pt}
        SPR & \textbf{82.9} & \textbf{79.6} & \textbf{64.8} & \textbf{76.5} & \textbf{69.4} & \textbf{56.0}  \\
        \bottomrule
    \end{tabular}
    \vskip 0.1in
    \caption{\textbf{Filtered noisy label percentages} in the purified buffer. 
    We compare SPR to two other state-of-the-art label filtering methods on CIFAR-100. We set $E_{max}=5$.
    The buffer size is set to 5000. ``random symmetric" refers to noise randomized across the 100 classes, while ``superclass symmetric" refers to noise randomized within the CIFAR-100 superclasses~\cite{krizhevsky09, lee20iclr}.}
    \label{tab:exp_cifar100_filter}
\end{table}

\begin{figure*}[t]
\begin{center}
    \includegraphics[width=\textwidth]{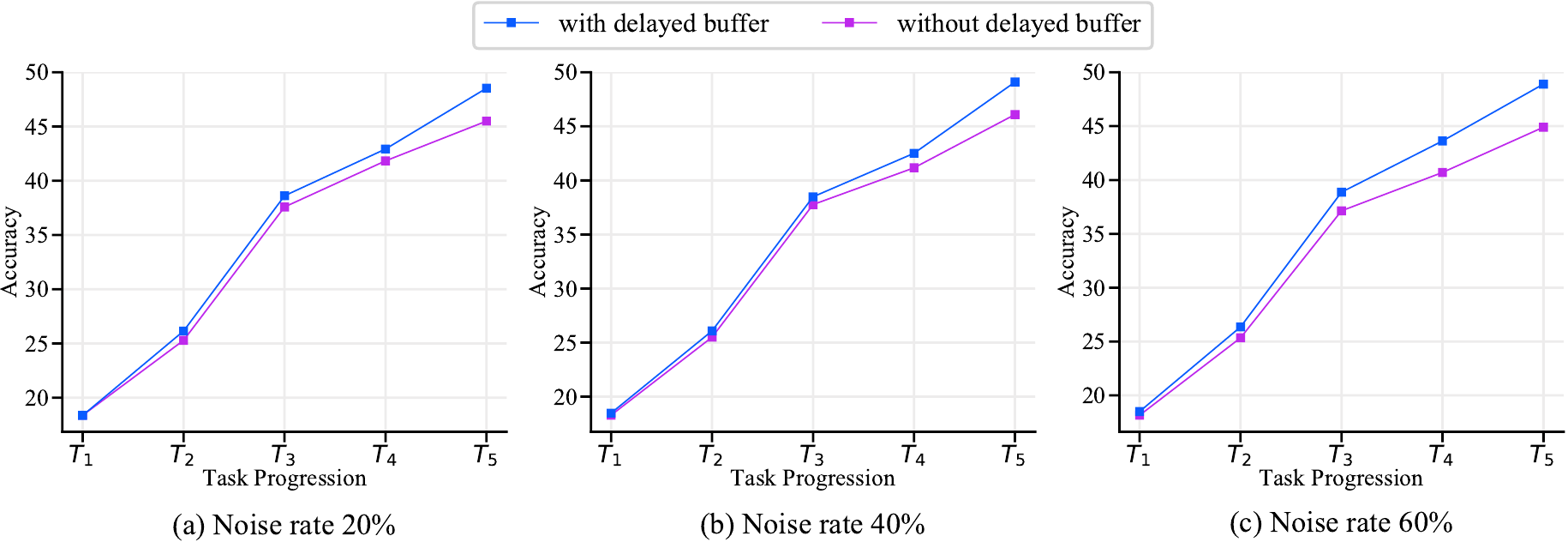}
\end{center}
\caption{The overall accuracy of SPR over sequential task progression on CIFAR-10 with different noise rates.
  Training with the delay buffer means that self-supervised learning is performed using the samples in both the delay buffer and the purified buffer,
  whereas training without the delay buffer means it is done with the samples in the purified buffer only.}
\label{fig:infoloss}
\end{figure*}
\subsection{Self-Replay with Noisy Labeled Data}
\label{sec:data_efficiency}

Table~\ref{tab:exp_infoloss} compares the overall accuracy of Self-Replay when self-supervised training is performed with and without the delay buffer.
Training with the delay buffer means using the samples in both the delay buffer $B_d$ (red) and the purified buffer $B_p$ (blue).
In contrast, training without the delay buffer means using purified samples $B_p$ (blue) only.
We remind the normalized temperature-scaled cross-entropy loss in the main manuscript as
\begin{align}
\label{eq:ntxent}
L_{self} =
- \hspace{-6pt}\sum_{i=1}^{2(\textcolor{red}{B_d} + \textcolor{blue}{B_p})} \hspace{-6pt}\text{log} \frac{e^{u_i^T u_j / \tau }}{\sum_{k=1}^{2(\textcolor{red}{B_d} + \textcolor{blue}{B_p})} \mathds{1}_{k \neq i} e^{u_i^T u_k / \tau}}. %
\end{align}

We observe an approximately  0.6\% increase in MNIST and 3.3\% increase in CIFAR-10 when using the delay buffer as well, even though it contains noisy labeled samples.
We speculate that slight improvement is attained in MNIST due to the simplicity of the features.
On the other hand, noticeable margins are seen in CIFAR-10, which we further analyze on a per-task basis, shown in Figure~\ref{fig:infoloss}.
The gaps are small in the earlier tasks but become more prominent as more tasks are seen. Moreover, the differences are even more significant when the level of noise rate increases.
The take-home message here is that self-supervised training can benefit from the increased data even if it could possibly contain noisy labels.·

\begin{table}[h]
    \centering
    \setlength{\tabcolsep}{5pt}
    \begin{tabular}{lccc|ccc}
        \toprule
        &\multicolumn{3}{c}{MNIST}  &\multicolumn{3}{c}{CIFAR-10} \\
        &\multicolumn{3}{c|}{symmetric}  &\multicolumn{3}{c}{symmetric} \\
        noise rate (\%)  & 20 & 40 & 60  & 20 & 40 & 60\\
        \midrule
        SR with DB & 91.0 & 91.8 & 91.1 & 48.5 & 49.1 & 48.9 \\
        \specialrule{0.1pt}{1pt}{1pt}
        SR without DB & 90.3 & 91.0 & 90.5 & 45.5 & 46.1 & 44.9 \\
        \bottomrule
    \end{tabular}
    \vskip 0.1in
    \caption{The overall accuracy of SPR with or without the samples in the delay buffer (DB).
      Self-supervised training can more benefit from more data even though some of them are possibly noisy·}
    \label{tab:exp_infoloss}
\end{table}

\subsection{Analyses of the Results on WebVision}
\label{sec:ext_ablation_study}
In the main manuscript, we briefly discuss the observation that Self-Replay and the Self-Centered filter do not synergize well on the WebVision dataset.
In this section, we provide extended discussions about this behavior with qualitative and quantitative analyses.

\begin{figure*}[h]
\begin{center}
    \includegraphics[width=\textwidth]{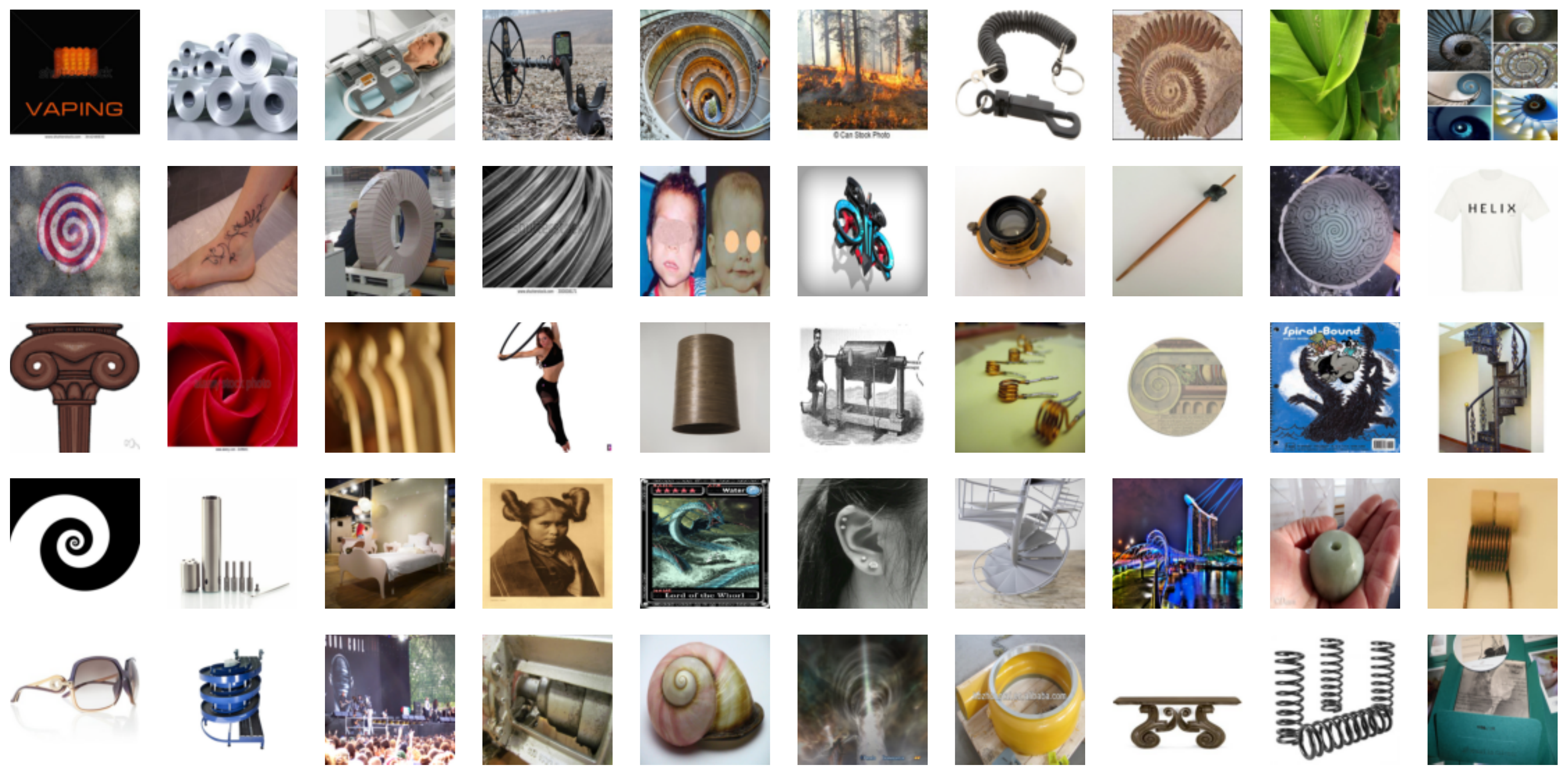}
\end{center}
\caption{50 random samples of the ``Spiral" class from the training set.}
\label{fig:spiral_random}
\end{figure*}

\begin{figure*}[h]
\begin{center}
    \includegraphics[width=\textwidth]{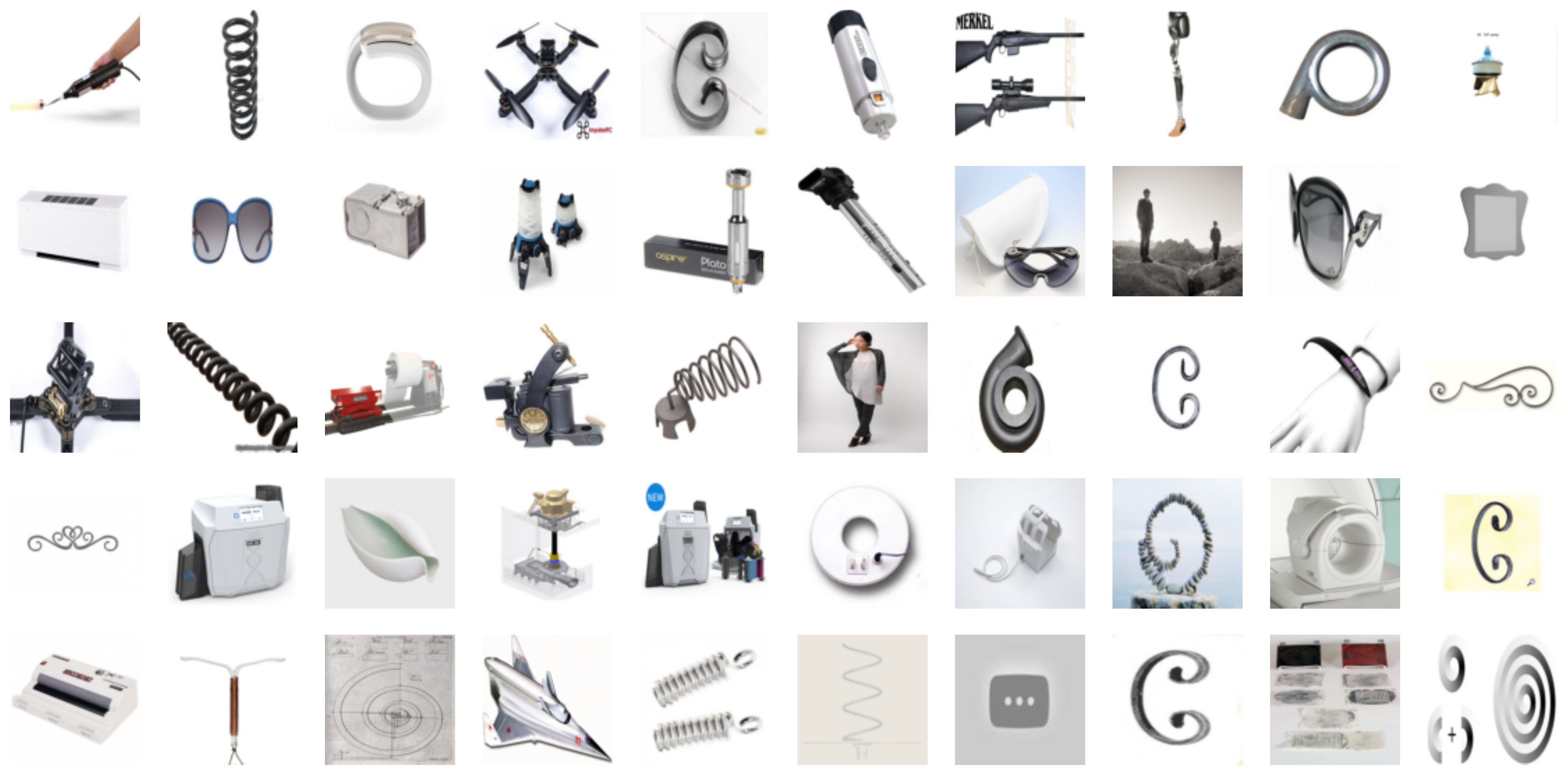}
\end{center}
\caption{50 random training samples of the ``Spiral" class from the purified buffer.}
\label{fig:spiral_filtered}
\end{figure*}

\begin{figure*}[h]
\begin{center}
    \includegraphics[width=\textwidth]{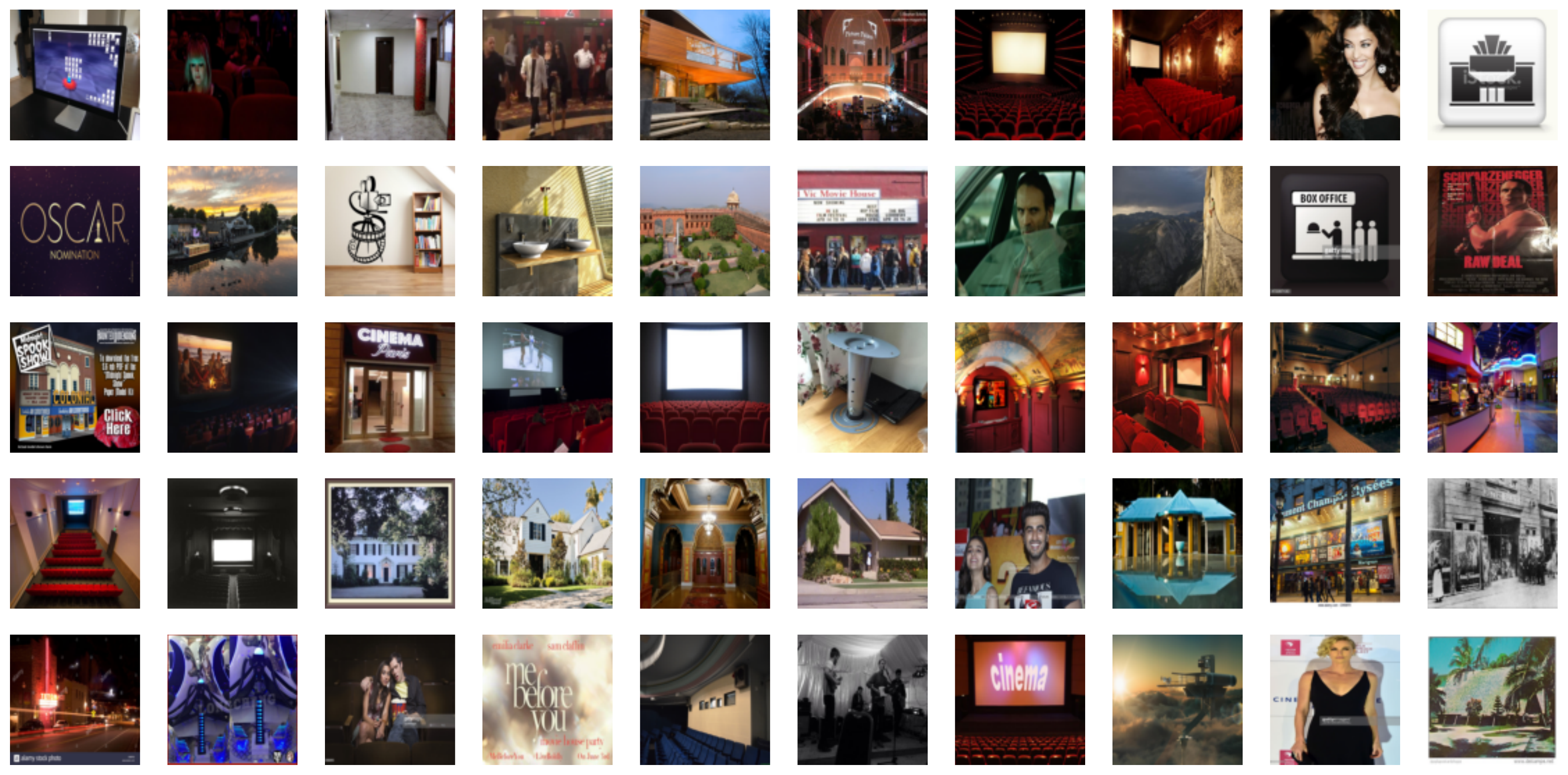}
\end{center}
\caption{50 random samples of the ``Cinema" class from the training set.}
\label{fig:cinema_random}
\end{figure*}

\begin{figure*}[h]
\begin{center}
    \includegraphics[width=\textwidth]{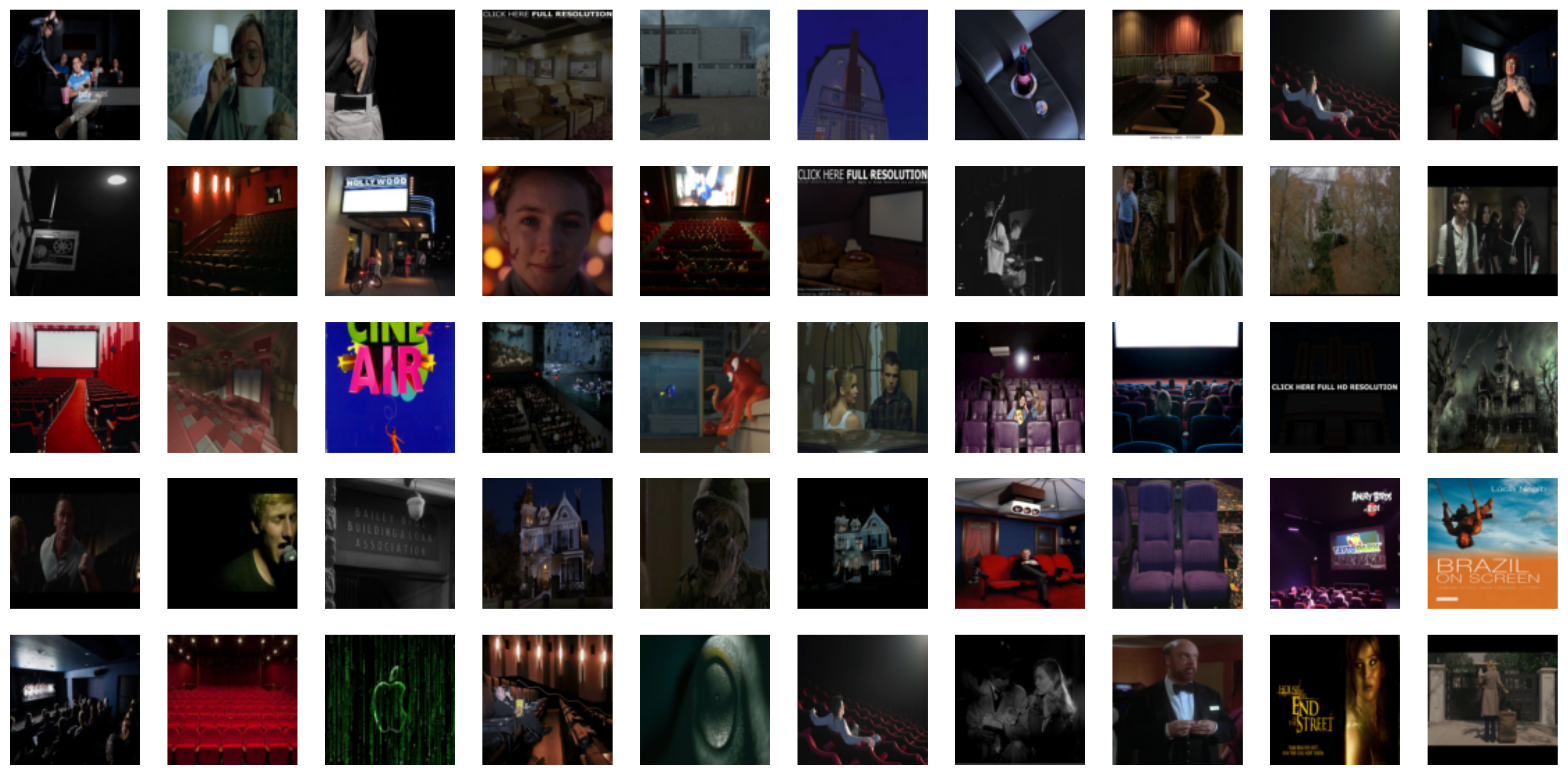}
\end{center}
\caption{50 random training samples of the ``Cinema" class from the purified buffer.}
\label{fig:cinema_filtered}
\end{figure*}

\textbf{Qualitative Analysis.} We pointed out that classes such as ``Spiral'' or ``Cinema'' are highly abstract by overarching broad related knowledge, which is at the same time corrupted by noise.
We show 50 random training data in Figure~\ref{fig:spiral_random} and Figure~\ref{fig:cinema_random} for ``Spiral'' and ``Cinema'', respectively.·
The Self-Centered filter samples for the same classes are also shown in Figure~\ref{fig:spiral_filtered} and Figure~\ref{fig:cinema_filtered}.
As visualized, it is not easy to interpret what the central concept is.

\begin{figure*}[h]
\begin{center}
    \includegraphics[width=\textwidth]{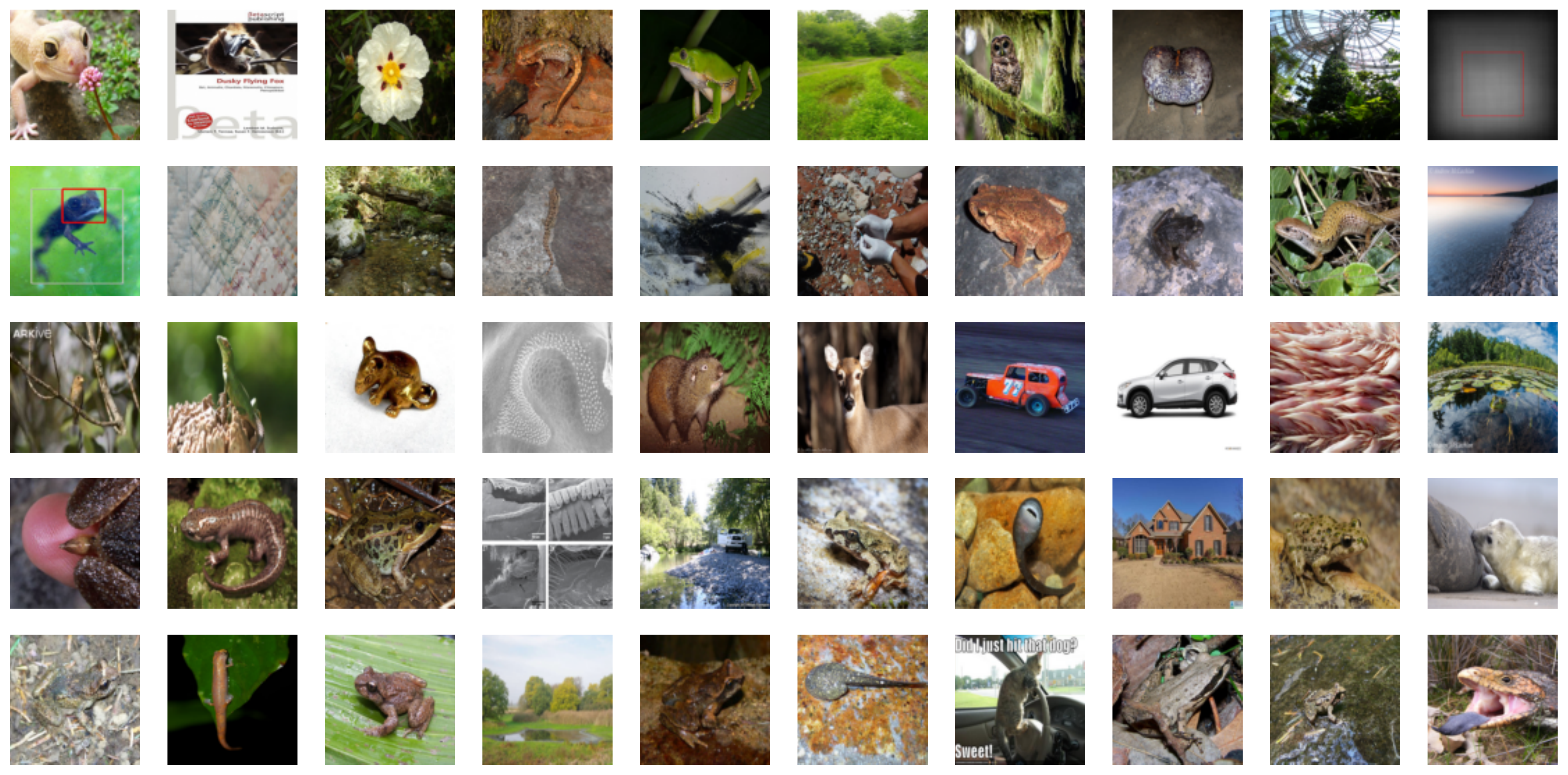}
\end{center}
\caption{50 samples of the ``Frog" class from the training set.}
\label{fig:frog_random}
\end{figure*}

\begin{figure*}[h]
\begin{center}
    \includegraphics[width=\textwidth]{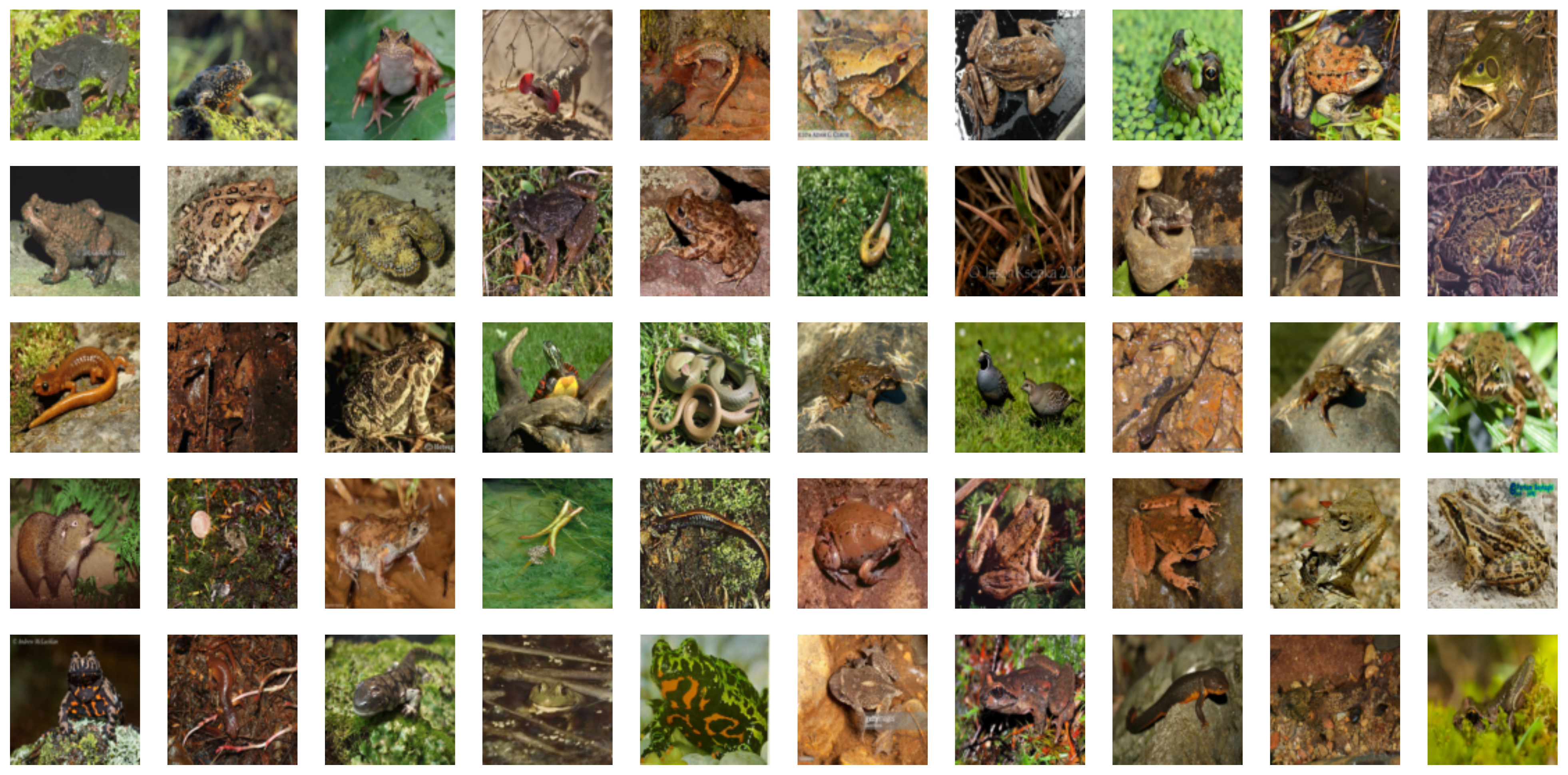}
\end{center}
\caption{50 training samples of the ``Frog" class from the purified buffer.}
\label{fig:frog_filtered}
\end{figure*}

\begin{figure*}[h]
\begin{center}
    \includegraphics[width=\textwidth]{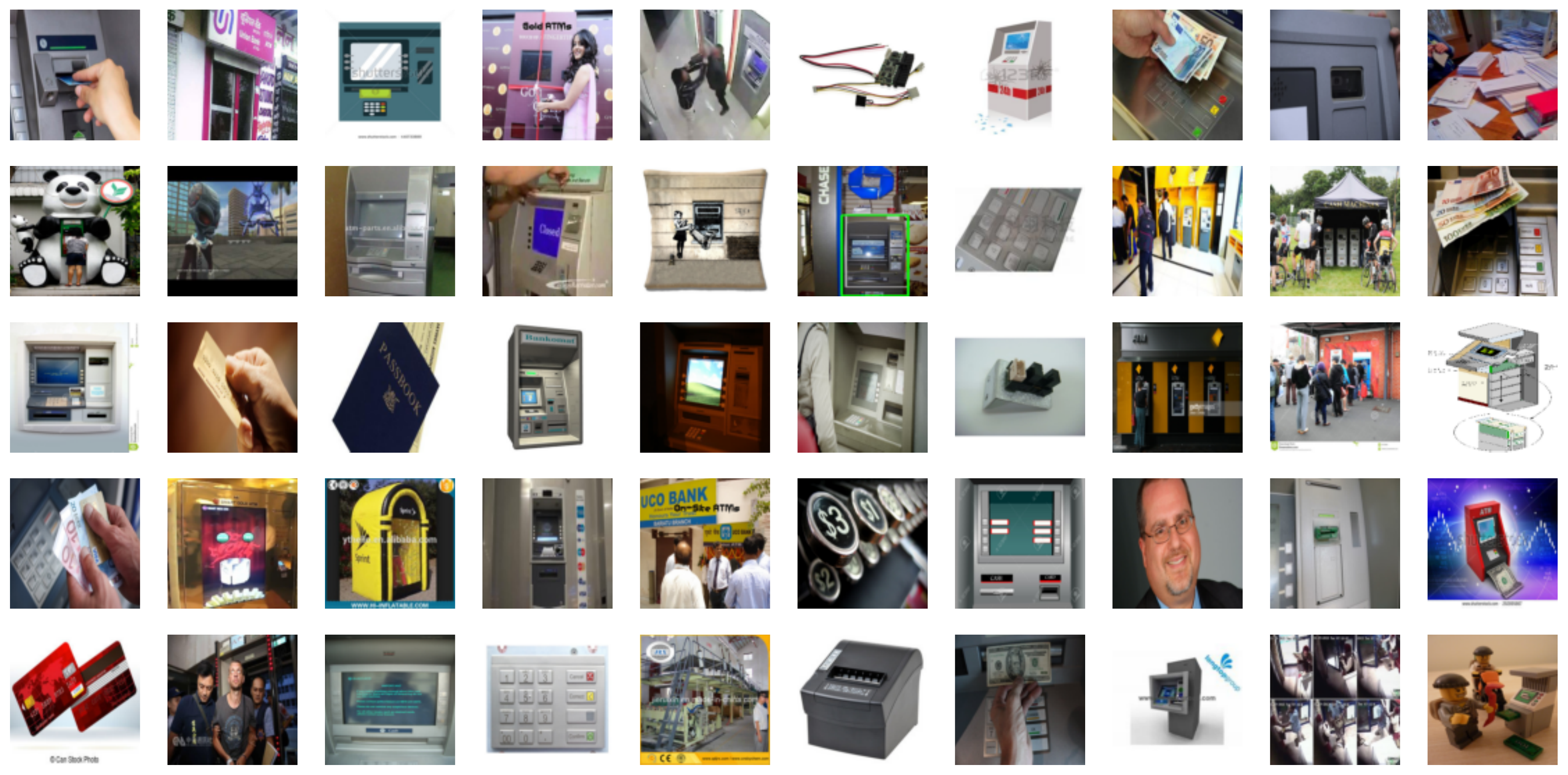}
\end{center}
\caption{50 samples of the ``ATM" class from the training set}
\label{fig:atm_random}
\end{figure*}

\begin{figure*}[h]
\begin{center}
    \includegraphics[width=\textwidth]{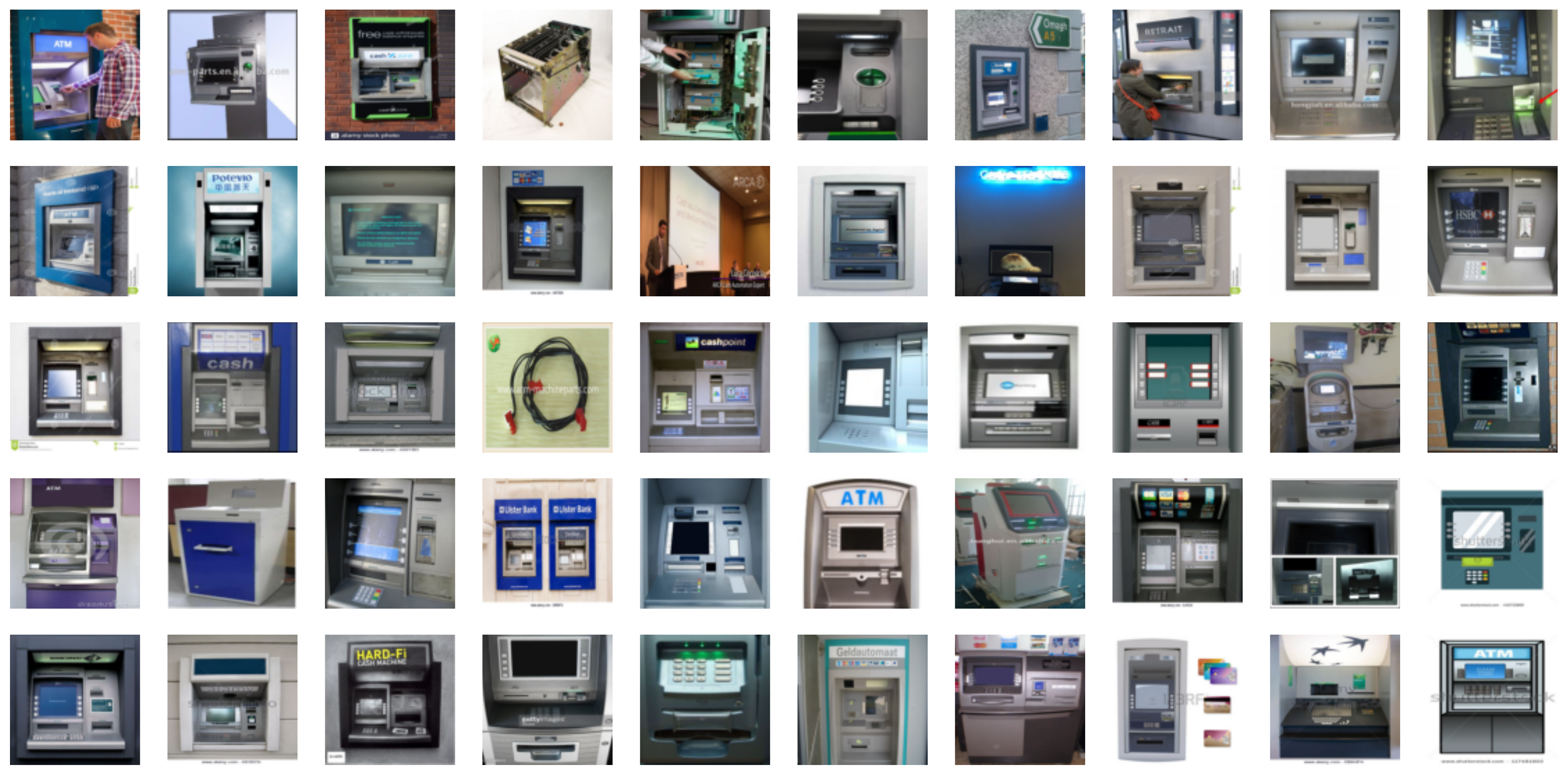}
\end{center}
\caption{50 random training samples of the ``ATM" class from the purified buffer.}
\label{fig:atm_filtered}
\end{figure*}

This is contrasted by the training samples in the classes ``ATM'' and ``Frog'' in Figure~\ref{fig:atm_random} and Figure~\ref{fig:frog_random}.
The classes contain noisy samples but represent the class concept without a high amount of abstraction.
We also show the Self-Centered filter samples for the classes in Figure~\ref{fig:atm_filtered} and Figure~\ref{fig:frog_filtered}.
It is much more visually evident what class the samples represent.

\textbf{Quantitative Analysis.} Table~\ref{tab:ext_ablation} contrasts the performance of the two topics on GDumb, Self-Replay, Self-Centered filter, and SPR.
The Self-Centered filter and SPR use the proposed Self-Centered filtering technique, whereas GDumb and Self-Replay use random sampling instead. 
The performances also support that random sampling may be a better performer for noisy and abstract classes, as GDumb and Self-Replay attain better performances.
On the other hand, for ordinary noisy classes such as ``ATM" or ``Frog," the Self-Centered filter and SPR perform stronger than random sampling and show a synergetic effect.

\begin{table}[t]
    \centering
    \small
    \setlength{\tabcolsep}{5pt}
    \begin{tabular}{lcccc}
        \toprule
               & GDumb & Self-Replay & Self-Centered filter & SPR \\
        \midrule
        ``Cinema" & 34.3  & 46.4        & 19.6                 & 26.8  \\
        \specialrule{0.1pt}{1pt}{1pt}
        ``Spiral" & 8.6   & 23.2        & 4.8                  & 9.0  \\
        \specialrule{0.4pt}{1pt}{1pt}
        \specialrule{0.4pt}{1pt}{1pt}
        ``ATM"    & 23.6  & 52.8        & 26.5                 & 54.0  \\
        \specialrule{0.1pt}{1pt}{1pt}
        ``Frog"    & 33.0   & 52.4        & 45.2                  & 55.0  \\

        \bottomrule
    \end{tabular}
    \vskip 0.1in
    \caption{Comparison of random sampling based methods (GDumb and Self-Replay) and the methods using the proposed Self-Centered filtering technique (Self-Centered filter and SPR). Random sampling is better for abstract classes such as ``Cinema" and ``Spiral", whereas Self-Centered filtering is better for ordinary noisy classes such as ``ATM" or ``Frog". The results are the mean of five unique random seed experiments.}
    \label{tab:ext_ablation}
\end{table}

\subsection{Episode Robustness}
\label{sec:episode_robust}
Table~\ref{tab:exp_accuracy_episodeB} (episode B) and Table~\ref{tab:exp_accuracy_episodeC} (episode C) report the results of two different randomly permuted episodes.
We include all of GDumb~\cite{prabhu19gdumb} combinations and the single best performing combination of PRS~\cite{kim20eccv} and CRS~\cite{vitter85} for each dataset.
Even in two additional random episode experiments, SPR performs much stronger than all the baselines on all datasets with real, symmetric, or asymmetric noise.

\begin{table*}[t]
    \centering
    \small
    \begin{tabular}{lccc|cc|ccc|cc|c}
        \toprule
        &\multicolumn{5}{c}{MNIST}         &\multicolumn{5}{c}{CIFAR-10} &\multicolumn{1}{c}{WebVision}\\
        &\multicolumn{3}{c|}{symmetric}    &\multicolumn{2}{c|}{asymmetric} &\multicolumn{3}{c|}{symmetric} &\multicolumn{2}{c|}{asymmetric} &\multicolumn{1}{c} {real noise}\\
        noise rate (\%)  & 20 & 40 & 60 & 20 & 40 & 20 & 40 & 60  & 20 & 40 & unknown \\
        \midrule
        Multitask 0\% noise~\cite{caruaca97} & \multicolumn{5}{c|}{98.6} &\multicolumn{5}{c|}{84.7} &\multicolumn{1}{c}{-}  \\
        Multitask~\cite{caruaca97} & 94.5  & 90.5  & 79.8 & 93.4 & 81.1 & 65.6 & 46.7 & 30.0 & 77.0 & 68.7 & 55.5 \\
        \specialrule{0.1pt}{1pt}{1pt}
        CRS + L2R~\cite{ren18l2r} & 80.8 & 74.1 & 59.7 & 85.3 & 79.8 & 29.8 & 23.1 & 16.0 & 36.4 & 36.1 & - \\
        CRS + Pencil~\cite{yi19pencil} & - & -  & - & - & - & - & - & - & - & - & 25.1 \\

        \specialrule{0.1pt}{1pt}{1pt}
        PRS + L2R~\cite{ren18l2r} & 80.7 & 74.0 & 60.4 & 83.2 & 80.1 & 30.8 & 22.8 & 15.0 & 36.3 & 32.9 & -  \\
        PRS + Pencil~\cite{yi19pencil} & - & - & - & - & - & - & - & - & - & - & 26.5 \\

        \specialrule{0.1pt}{1pt}{1pt}
        MIR + L2R~\cite{ren18l2r} & 79.6 & 68.6 & 51.6 & 83.2 & 79.5 & 31.1 & 21.0 & 14.5 & 34.7 & 33.6 & -  \\
        MIR + Pencil~\cite{yi19pencil} & - & - & - & - & - & - & - & - & - & - & 22.6  \\

        \specialrule{0.1pt}{1pt}{1pt}
        GDumb~\cite{prabhu19gdumb}  & 70.1 & 54.6 & 32.3 & 78.2 & 71.1 & 29.6 & 22.4 & 16.5 & 33.0 & 30.9 & 33.3  \\
        GDumb + L2R~\cite{ren18l2r} & 67.1 & 59.2 & 40.6 & 70.6 & 68.7 & 27.0 & 25.5 & 21.8 & 29.9 & 29.4 & - \\
        GDumb + Pencil~\cite{yi19pencil} & 70.2 & 53.9 & 35.4 & 77.5 & 70.2 & 28.1 & 21.0 & 15.9 & 31.5 & 30.6 & 27.5  \\
        GDumb + SL~\cite{wang19sce} & 65.6 & 47.5 & 30.5 & 73.3 & 68.5 & 27.1 & 22.6 & 16.8 & 33.2 & 31.4 & 32.5  \\
        GDumb + JoCoR~\cite{wei20jocor} & 68.3 & 56.0 & 41.0 & 78.5 & 70.9 & 26.6 & 21.1 & 15.9 & 32.9 & 32.2 & 22.9  \\

        \specialrule{0.4pt}{1pt}{1pt}
        SPR            & \textbf{86.8}          & \textbf{87.2}          & \textbf{82.1}          & \textbf{86.6}                  & \textbf{85.5}            & \textbf{42.0}          & \textbf{42.4}          & \textbf{39.1}          & \textbf{44.4}          & \textbf{43.3}         &  \textbf{41.6} \\

        \bottomrule
    \end{tabular}

    \vskip 0.1in
    \caption{\textbf{Overall accuracy on episode B} after all sequences of tasks are trained. The buffer size is set to 300, 500, 1000 for MNIST, CIFAR-10, and WebVision, respectively.
        We report all of GDumb~\cite{prabhu19gdumb} combinations and single best performing combination of PRS~\cite{kim20eccv} and CRS~\cite{vitter85}.
        Some empty slots on WebVision are due to the unavailability of clean samples required by L2R for training~\cite{ren18l2r}.
        The results are the mean of five unique random seed experiments.}
    \label{tab:exp_accuracy_episodeB}
\end{table*}

\begin{table*}[t]
    \centering
    \small
    \begin{tabular}{lccc|cc|ccc|cc|c}
        \toprule
        &\multicolumn{5}{c}{MNIST}         &\multicolumn{5}{c}{CIFAR-10} &\multicolumn{1}{c}{WebVision}\\
        &\multicolumn{3}{c|}{symmetric}    &\multicolumn{2}{c|}{asymmetric} &\multicolumn{3}{c|}{symmetric} &\multicolumn{2}{c|}{asymmetric} &\multicolumn{1}{c} {real noise}\\
        noise rate (\%)  & 20 & 40 & 60 & 20 & 40 & 20 & 40 & 60  & 20 & 40 & unknown \\
        \midrule
        Multitask 0\% noise~\cite{caruaca97} & \multicolumn{5}{c|}{98.6} &\multicolumn{5}{c|}{84.7} &\multicolumn{1}{c}{-}  \\
        Multitask~\cite{caruaca97} & 94.5  & 90.5  & 79.8 & 93.4 & 81.1 & 65.6 & 46.7 & 30.0 & 77.0 & 68.7 & 55.5 \\
        \specialrule{0.1pt}{1pt}{1pt}
        CRS + L2R~\cite{ren18l2r} & 79.9 & 74.9 & 58.2 & 84.4 & 79.4 & 29.3 & 24.4 & 16.8 & 37.2 & 37.5 & - \\
        CRS + Pencil~\cite{yi19pencil} & - & -  & - & - & - & - & - & - & - & - & 29.9 \\

        \specialrule{0.1pt}{1pt}{1pt}
        PRS + L2R~\cite{ren18l2r} & 80.5 & 72.3 & 55.2 & 83.8 & 80.1 & 30.6 & 23.3 & 16.3 & 37.2 & 36.1 & -  \\
        PRS + Pencil~\cite{yi19pencil} & - & - & - & - & - & - & - & - & - & - & 28.5 \\

        \specialrule{0.1pt}{1pt}{1pt}
        MIR + L2R~\cite{ren18l2r} & 80.3 & 69.7 & 47.1 & 83.0 & 77.6 & 28.2 & 21.3 & 15.6 & 36.3 & 34.3 & -  \\
        MIR + Pencil~\cite{yi19pencil} & - & - & - & - & - & - & - & - & - & - & 22.4  \\

        \specialrule{0.1pt}{1pt}{1pt}
        GDumb~\cite{prabhu19gdumb}  & 71.8 & 52.8 & 37.5 & 79.2 & 72.1 & 28.7 & 23.0 & 16.3 & 34.2 & 31.9 & 31.6  \\
        GDumb + L2R~\cite{ren18l2r} & 67.7 & 58.2 & 42.7 & 69.3 & 67.6 & 28.9 & 24.8 & 19.7 & 31.8 & 29.4 & - \\
        GDumb + Pencil~\cite{yi19pencil} & 69.0 & 54.2 & 37.8 & 78.6 & 71.2 & 27.5 & 21.0 & 16.6 & 31.3 & 31.8 & 28.5  \\
        GDumb + SL~\cite{wang19sce} & 65.4 & 48.4 & 29.1 & 72.4 & 67.7 & 28.3 & 22.9 & 15.0 & 31.4 & 31.9 & 31.6  \\
        GDumb + JoCoR~\cite{wei20jocor} & 70.4 & 59.0 & 40.6 & 77.4 & 70.6 & 27.8 & 22.3 & 15.5 & 33.4 & 31.7 & 24.3  \\

        \specialrule{0.4pt}{1pt}{1pt}
        SPR            & \textbf{86.6}          & \textbf{87.5}          & \textbf{84.4}          & \textbf{87.0}                  & \textbf{87.3}            & \textbf{43.7}          & \textbf{43.1}          & \textbf{39.8}          & \textbf{44.3}          & \textbf{43.2}         &  \textbf{40.2} \\

        \bottomrule
    \end{tabular}

    \vskip 0.1in
    \caption{\textbf{Overall accuracy on episode C} after all sequences of tasks are trained. The buffer size is set to 300, 500, 1000 for MNIST, CIFAR-10, and WebVision, respectively.
        We report all of GDumb~\cite{prabhu19gdumb} combinations and single best performing combination of PRS~\cite{kim20eccv} and CRS~\cite{vitter85}.
        Some empty slots on WebVision are due to the unavailability of clean samples required by L2R for training~\cite{ren18l2r}.
        The results are the mean of five unique random seed experiments.}
    \label{tab:exp_accuracy_episodeC}
\end{table*}

\subsection{Buffer Size Analysis}
\label{sec:buffsize_analysis}
SPR requires a larger amount of memory than some baselines (excluding L2R), but the usage of the memory is different in that, 
a hold-out memory (Delay Buffer) is used for the purpose of filtering out the noisy labels, while only the Purified Buffer is used to mitigate the amount of forgetting.
Hence, simply giving the other baselines a replay buffer twice as big would not be a fair comparison in the viewpoint of continual learning alone. 
Nonetheless, we run the experiments shown in Table~\ref{tab:exp_accuracy_doublebuff}, where all of GDumb~\cite{prabhu19gdumb} combinations are allowed twice the buffer size for replay. 
Even so, SPR using half the buffer size is able to outperform all the other baselines. 
Furthermore, to inform how the buffer size affects the results, we halve the original used buffer size and report the results in Table~\ref{tab:exp_accuracy_halfbuff}. 
SPR still strongly outperforms the baselines in all the datasets and noise rates.
These two experiments show that SPR is robust to the buffer size, and its performance is due to self-supervised learning and the clean-buffer management, rather than using the hold-out memory for the Delay buffer.

\begin{table*}[t]
    \centering
    \small
    \begin{tabular}{lccc|cc|ccc|cc|c}
        \toprule
        &\multicolumn{5}{c}{MNIST}         &\multicolumn{5}{c}{CIFAR-10} &\multicolumn{1}{c}{WebVision}\\
        &\multicolumn{3}{c|}{symmetric}    &\multicolumn{2}{c|}{asymmetric} &\multicolumn{3}{c|}{symmetric} &\multicolumn{2}{c|}{asymmetric} &\multicolumn{1}{c} {real noise}\\
        Buffer size  & \multicolumn{3}{c|}{150}& \multicolumn{2}{c|}{150}& \multicolumn{3}{c|}{250} & \multicolumn{2}{c|}{250}& \multicolumn{1}{c}{500}\\
        noise rate (\%)  & 20 & 40 & 60 & 20 & 40 & 20 & 40 & 60  & 20 & 40 & unknown \\
        \midrule
        GDumb + L2R~\cite{ren18l2r} & 64.8 & 55.5 & 37.8 & 71.2 & 66.8 & 23.2 & 22.1 & 19.3 & 28.4 & 24.8 & - \\
        GDumb + Pencil~\cite{yi19pencil} & 59.3 & 48.1 & 36.4 & 76.4 & 66.6 & 25.6 & 17.9 & 13.9 & 27.6 & 26.8 & 21.1  \\
        GDumb + SL~\cite{wang19sce} & 61.5 & 41.3 & 31.1 & 66.8 & 56.8 & 20.7 & 19.8 & 18.8 & 29.2 & 26.4 & 26.4  \\
        GDumb + JoCoR~\cite{wei20jocor} & 66.8 & 60.9 & 33.0 & 74.4 & 66.3 & 23.8 & 18.9 & 14.2 & 26.2 & 26.2 & 23.0  \\

        \specialrule{0.4pt}{1pt}{1pt}
        SPR            & \textbf{82.6}          & \textbf{85.4}          & \textbf{81.2}          & \textbf{77.0}                  & \textbf{81.6}            & \textbf{41.2}          & \textbf{41.2}          & \textbf{37.8}          & \textbf{42.8}          & \textbf{41.3}         &  \textbf{39.4} \\

        \bottomrule
    \end{tabular}

    \vskip 0.1in
    \caption{\textbf{Overall accuracy on the half buffer size} after all sequences of tasks are trained. The buffer size is set to 150, 250, 500 for MNIST, CIFAR-10, and WebVision, respectively.
        We report all of GDumb~\cite{prabhu19gdumb} combinations. An empty slot on WebVision are due to the unavailability of clean samples required by L2R for training~\cite{ren18l2r}.}
    \label{tab:exp_accuracy_halfbuff}
\end{table*}

\begin{table*}[t]
    \centering
    \small
    \begin{tabular}{lccc|cc|ccc|cc|c}
        \toprule
        &\multicolumn{5}{c}{MNIST}         &\multicolumn{5}{c}{CIFAR-10} &\multicolumn{1}{c}{WebVision}\\
        &\multicolumn{3}{c|}{symmetric}    &\multicolumn{2}{c|}{asymmetric} &\multicolumn{3}{c|}{symmetric} &\multicolumn{2}{c|}{asymmetric} &\multicolumn{1}{c} {real noise}\\
        Buffer size  & \multicolumn{3}{c|}{600}& \multicolumn{2}{c|}{600}& \multicolumn{3}{c|}{1000} & \multicolumn{2}{c|}{1000}& \multicolumn{1}{c}{2000}\\
        noise rate (\%)  & 20 & 40 & 60 & 20 & 40 & 20 & 40 & 60  & 20 & 40 & unknown \\
        \midrule
        GDumb + L2R~\cite{ren18l2r} & 76.7 & 62.6 & 51.9 & 79.7 & 73.3 & 31.4 & 27.3 & 24.0 & 35.0 & 36.0 & - \\
        GDumb + Pencil~\cite{yi19pencil} & 72.1 & 58.5 & 39.4 & 75.3 & 73.5 & 31.2 & 24.5 & 16.4 & 38.6 & 35.5 & 33.0  \\
        GDumb + SL~\cite{wang19sce} & 66.0 & 47.2 & 31.7 & 79.0 & 74.8 & 33.1 & 23.2 & 17.7 & 40.4 & 37.3 & 38.5  \\
        GDumb + JoCoR~\cite{wei20jocor} & 74.3 & 57.8 & 42.5 & 78.3 & 76.0 & 31.9 & 22.8 & 17.4 & 42.5 & 38.1 & 27.0  \\

        \midrule 
        \midrule 
        Buffer size  & \multicolumn{3}{c|}{300}& \multicolumn{2}{c|}{300}& \multicolumn{3}{c|}{500} & \multicolumn{2}{c|}{500}& \multicolumn{1}{c}{1000}\\
        \specialrule{0.4pt}{1pt}{1pt}
        SPR            & \textbf{85.4}          & \textbf{86.7}          & \textbf{84.8}          & \textbf{86.8}                  & \textbf{86.0}            & \textbf{43.9}          & \textbf{43.0}          & \textbf{40.0}          & \textbf{44.5}          & \textbf{43.9}         &  \textbf{40.0} \\

        \bottomrule
    \end{tabular}

    \vskip 0.1in
    \caption{\textbf{Overall accuracy on the double buffer size for all of GDumb combinations} after all sequences of tasks are trained. The buffer size is set to 600, 1000, 2000 for MNIST, CIFAR-10, and WebVision, respectively.
        An empty slot on WebVision are due to the unavailability of clean samples required by L2R for training~\cite{ren18l2r}.
        Note that SPR outperforms all of GDumb~\cite{prabhu19gdumb} combinations with the buffer size of 300, 500, 1000 for MNIST, CIFAR-10, and WebVision, respectively.
    }
    \label{tab:exp_accuracy_doublebuff}
\end{table*}

\subsection{Variance}
\label{sec:variance}
\begin{figure}[ht]
\begin{center}
    \includegraphics[width=\columnwidth]{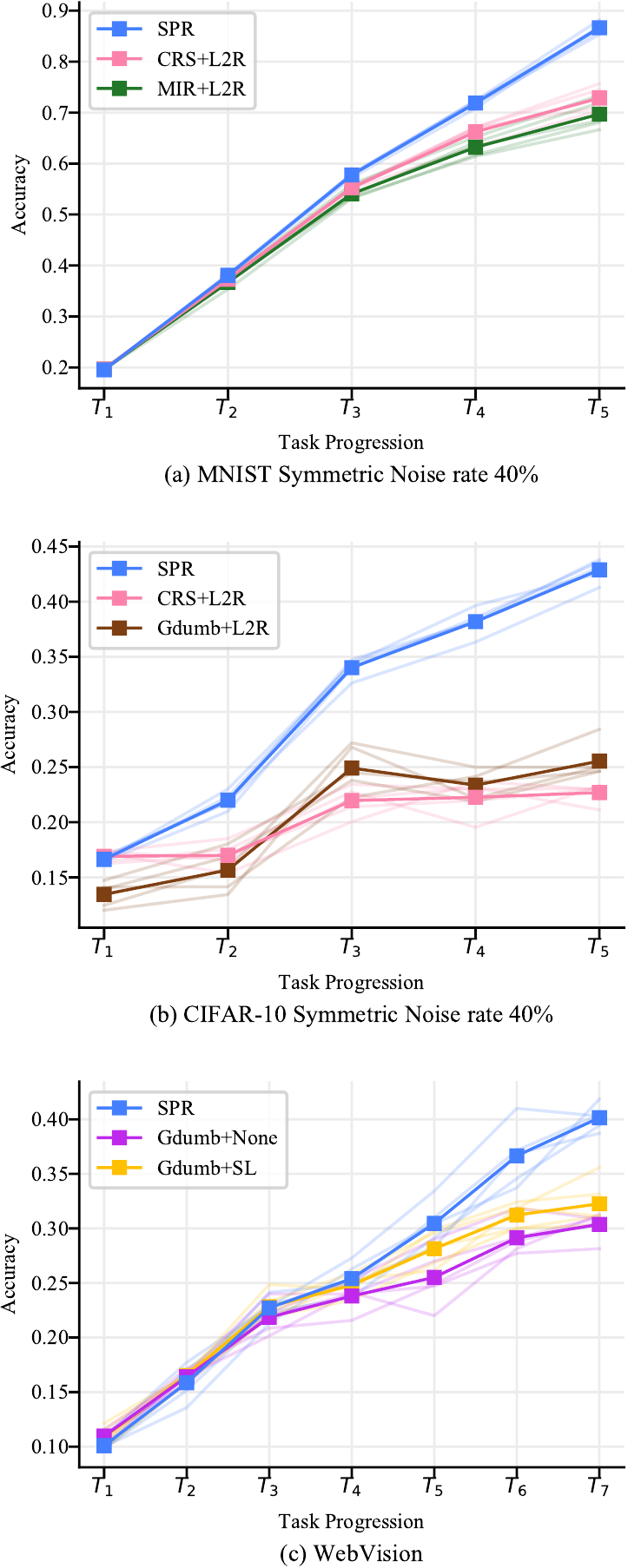}
\end{center}
\caption{Accuracy and variances of top-3 best-performing methods for MNIST, CIFAR-10 and WebVision.}
\label{fig:top3_variance}
\end{figure}

Figure~\ref{fig:top3_variance} visualizes the variances of top-3 best-performing methods for MNIST, CIFAR-10 with 40\% symmetric noise rate, and WebVision with real-noise.
Among the symmetric noise experiments with five different random seeds, SPR shows a minor amount of variance throughout the tasks. However, for WebVision, a noticeable amount of fluctuations are seen for all three approaches.

\end{appendix}

\end{document}